\documentclass[titlepage, 12pt]{article}
\usepackage[utf8]{inputenc}
\usepackage{amssymb, amsmath, mathtools, bm}
\usepackage[font=small]{caption}
\usepackage{geometry}
\usepackage{natbib}
\usepackage{todonotes}
\usepackage{csquotes}
\usepackage{graphicx}
\usepackage{wrapfig}
\usepackage{booktabs}
\usepackage[hidelinks]{hyperref}
\usepackage{url}

\usepackage{fourier}
\setcitestyle{authoryear, open={(},close={)}}

\title{%
Argument Linking \\
\large a survey and forecast}
\author{William A. H. Gantt IV}
\date{Spring 2021}
\begin{document}

\maketitle

\begin{abstract}
    Semantic role labeling (SRL)---identifying the semantic relationships between a predicate and other constituents in the same sentence---is a well-studied task in natural language understanding (NLU). However, many of these relationships are evident only at the level of the document, as a role for a predicate in one sentence may often be filled by an argument in a different one. This more general task, known as \textit{implicit semantic role labeling} or \textit{argument linking}, has received increased attention in recent years, as researchers have recognized its centrality to information extraction and NLU.\@ This paper surveys the literature on argument linking and identifies several notable shortcomings of existing approaches that indicate the paths along which future research effort could most profitably be spent.
\end{abstract}

\tableofcontents

\section{Introduction}
\label{sec:introduction}
Semantic role labeling (SRL) is the task of identifying the semantic relationships between a predicate and its syntactic constituents by assigning those constituents to the appropriate role(s) evoked by the predicate. SRL's salience to general natural language understanding (NLU) hardly requires justification: the better part of understanding the meaning of a sentence just \textit{is} grasping the semantic relationships between an event (described by a predicate) and its participants (i.e.\ its arguments). Since the first systems for automated SRL were introduced by \citet{gildea2002automatic}, efforts to improve machine performance at the task have multiplied, and they have largely followed the field more broadly in moving from heavy feature engineering and classic ML algorithms \citep{xue-palmer-2004-calibrating, pradhan2005support, koomen2005generalized} to deep learning methods more recently \citep{lee-etal-2017-end, he-etal-2017-deep, peters-etal-2018-deep}.

The focus in SRL has, however, remained overwhelmingly at the sentence level. This is not without some reason. For one, the majority of a predicate's semantic roles are filled by sentence-local arguments \citep{ebner-etal-2020-multi}. For another, the more general problem of \textit{implicit semantic role labeling} (iSRL) or \textit{argument linking}---identifying \textit{all} explicit arguments that fill a predicate's roles, local or not---is decidedly harder. Where SRL systems have often relied on syntactic parses to aid in labeling, such information is far less helpful when considering cross-sentential relations. What's more, the number of candidate arguments for a particular role is plainly much larger when all sentences in the document must potentially be considered. And finally, unlike SRL, argument linking is unavoidably a coreference problem, as the same entities must fill the same roles of predicates describing identical events. This brings with it a burden of document-level coherence that traditional SRL has been able to ignore.

These differences with SRL also hint at argument linking's kinship with other major information extraction (IE) objectives---not only entity and event coreference, but the classic problems of template and slot filling, schema induction, and general document-level relation extraction. Each of these have vast literatures of their own and I do not attempt to review them here. Rather, I focus only on what I consider to be argument linking proper, coming into contact with these other objectives only insofar as they are directly relevant.

In the remainder of the paper, I present the history of argument linking up to the present, organized by the major datasets and tasks in which it has been formalized (\S\ref{sec:arg_linking}) and conclude by describing what I perceive to be the major gaps in the literature, along with some proposed bridges (\S\ref{sec:future_work}).

\section{Argument Linking: An Overview}
\label{sec:arg_linking}
\subsection{Terminology}
\label{subsec:terminology}
Before venturing into the literature on argument linking, it will be useful to define a number of important technical terms. First, I take an \textit{event trigger} to be a minimal span of text that refers to an event, where \textit{event} should be construed as broadly as possible, incorporating states of affairs (e.g.\ \textit{is sunny}) as well as occurrences that involve some change (e.g.\ \textit{run}). Event triggers may include not only verb phrases but also event-denoting nominals (e.g.\ \textit{the birthday party}). Frequently, I will follow previous authors in using \textit{predicate} in a sense that is, for practical purposes, synonymous with \textit{event trigger}.

Following \citet{ebner-etal-2020-multi}, I assume that an event trigger evokes a set of semantic roles that may or may not be filled by entities explicitly mentioned somewhere in the text. Explicitly mentioned entities that fill one or more roles for a given trigger are the \textit{explicit arguments} of the trigger, regardless of whether they appear in the same sentence. For instance, in the sentence \textit{Navin carried the chair}, both \textit{Navin} and \textit{the chair} are explicit arguments of \textit{carried}---perhaps filling the Agent and Patient roles, respectively, in a traditional SRL ontology.\footnote{The example is taken from \citep{williams2015arguments}} The carrying must also have happened at a particular time and in a particular place, but if this information isn't given in the text, a Time or Place role will have no explicit arguments.

Turning to semantic roles directly, I follow O'Gorman's (2019) definition of \textit{implicit roles} as ``instances where a linguistically mentioned event can be inferred to have a particular participant, but where there is no explicit syntactic encoding expressing that relationship to the participant.'' I will sometimes also use \textit{explicit role} to refer to semantic roles where this ``explicit syntactic encoding'' is present. On these definitions, then, explicit roles \textit{must} have explicit arguments and implicit roles \textit{can} have explicit arguments, but may not. 

If there are such things as explicit arguments, it seems that there should also be implicit ones. However, I have avoided the term \textit{implicit argument} entirely, owing to the thoroughly muddled usage it has across IE, NLP, and linguistics.\footnote{The meaning of \textit{explicit argument} is somewhat muddled as well, but I have found it harder to avoid, and the definition I gave above is reasonably straightforward.} In IE and NLP, an explicit argument is most commonly taken to be a constituent that fills one of the roles evoked by the trigger and that furthermore bears a direct syntactic relationship to it; an implicit argument is then any constituent that fills a role evoked by the trigger but does not participate in a syntactic relationship. Other times, the distinction is conflated with sentence locality: explicit arguments are role fillers that merely appear somewhere in the same sentence as the trigger; implicit ones are role fillers that do not. Still more confusingly, \citet{ebner-etal-2020-multi} seemingly use \textit{implicit argument} as synonymous with \textit{semantic role} generally, and \citet{williams2015arguments} proposes two senses of \textit{implicit argument} that are somehow both different from all of the above. When I have to make distinctions between kinds of explicit argument on the basis of sentence locality or syntactic dependency, I will simply use the terms \textit{local} or \textit{non-local} argument and \textit{syntactic} or \textit{non-syntactic} argument.

As suggested in \S\ref{sec:introduction}, I take \textit{implicit semantic role labeling} and \textit{argument linking} to be synonymous, both denoting the problem of identifying all of the explicit arguments of the roles evoked by the event triggers in a document.\footnote{For consistency, I arbitrarily use \textit{argument linking} throughout.} This definition is deliberately agnostic on several points. Most obviously, the definition is agnostic to the role ontology---the set of all roles that a trigger might evoke. Different resources present different ontologies and, more importantly, there is longstanding theoretical debate over the proper nature and number of roles \citep{dowty1991thematic, jackendoff1987status, wechsler1995semantic, williams2015arguments}. The definition also does not indicate whether gold argument spans are known; both settings will be on display in the tasks covered here.\footnote{See \S\ref{subsec:model_eval_problems} for a more partisan take on this matter.} Lastly, it makes no assumptions about the number of roles a single argument can fill or about the number of arguments a single role can have. There are compelling reasons to think one should \textit{not} be agnostic on the latter front---that the same role never binds more than one argument---but this introduces dependencies between roles and arguments that are often cumbersome to model computationally and that are accordingly ignored.\footnote{\citet{williams2015arguments}, for instance, presents a variety of arguments for what he calls the \textit{Role Iteration Generalization}---the thesis that ``generally or always, two distinct dependents do not bind the same type of semantic relation'' (163).}

Agnosticism to ontologies aside, argument linking, like SRL, has drawn especially heavily on the theory of frame semantics developed by Charles Fillmore and on the associated FrameNet project \citep{fillmore1976frame, baker1998berkeley}. Briefly, frame semantics holds that many word meanings are best understood in terms of the \textit{semantic frame} that they evoke, where a frame describes a (type of) event, relation, or entity, and a set of associated roles called \textit{frame elements} (FEs). Multiple words may evoke the same frame, and the same word may evoke different frames by virtue of polysemy. The set of words that evoke a given frame are known as its \textit{lexical units} (LUs). Frame semantics is relevant here principally for its notion of \textit{null instantiation} (NI), which refers to cases in which an FE is not realized (``instantiated'') by any entity in the text. NIs are traditionally divided into three kinds based on their interpretation:
\begin{enumerate}
    \item \textit{Definite} (DNI). The NI refers to a definite entity that is discernible from context, as in the \textit{Goal} relation in \textit{we arrived $[\emptyset_{Goal}]$}.
    \item \textit{Indefinite} (INI). The the NI refers to an absent object of a transitive verb, which receives an existential interpretation, as the \textit{Theme} relation in \textit{I was eating $[\emptyset_{Theme}]$}. 
    \item \textit{Constructional} (CNI). The NI is an artifact of a particular syntactic construction, and typically also receives an existential interpretation, as the \textit{Agent} relation in the passive sentence \textit{$[\emptyset_{Agent}]$ Mistakes were made}.
\end{enumerate}
Most work in argument linking has focused on resolving cases of DNI, but some tasks also feature a preliminary objective in which labeled NIs are given, but their type must be determined. I begin this survey with one such task, SemEval-2010 Task 10.

\subsection{SemEval-2010 Task 10}
\label{subsec:semeval-task-10}
SemEval 2010 Task 10 (``linking events and their participants in discourse''---henceforth just ``SemEval'') introduced the first dataset and suite of tasks for argument linking, and has been a standard evaluation benchmark since then \citep{ruppenhofer-etal-2010-semeval}. The dataset annotates the FrameNet frames for event triggers in excerpts from \textit{Sherlock Holmes} stories, along with all null instantiations (both DNIs and INIs) for the triggers' frame elements. Each DNI is associated with a coreference chain comprising all mentions of the DNI's referent in the document. In total, the dataset comprises just 963 sentences (438 train, 525 test), 652 instances of DNI (303 train, 349 test) and 638 instances of INI (277 train, 361 test). The text for the training set comes from two \textit{Sherlock Holmes} short stories (``The Adventure of Wisteria Lodge'' and ``The Tiger of San Pedro'') and the text for the test set comes from the final two chapters of \textit{The Hound of the Baskervilles}. 

As originally formulated, the task is divided into three settings, based on whether only traditional SRL is to be performed, only DNI referents are to be linked to their appropriate trigger and role, or both are to be done:

\begin{enumerate}
    \item \textbf{Full task.} Given gold frames for the triggers, perform both SRL and NI linking.
    \item \textbf{NI only.} Given both gold frames and labeled semantic roles, perform NI linking only.
    \item \textbf{SRL only.} Given gold frames, perform SRL only (no NI linking).
\end{enumerate}

Only three teams participated in the original evaluation, and of these, only one (\citet{chen2010semafor}, discussed below) attempted task 1. Since then, most efforts have focused only on task 2. For NI linking, gold argument spans are \textit{not} assumed. The official evaluation scoring in this setting is based on the \textit{Dice coefficient} between words in the predicted ($P_r$) and gold ($G_r$) argument spans for a particular role instance $r$:

$$score(P_r,G_r) = \frac{2 \cdot |P_r \cap G_r|}{|P_r| + |G_r|}$$

\noindent where ``$|\cdot|$'' denotes the length of the relevant span and ``$\cap$'' denotes token-level overlap. This score has the advantage of dissuading competing systems from gaming the objective by simply selecting very large spans for linking. Given that SemEval provides coreference chains, there are potentially multiple correct arguments. Thus, the max coefficient across all gold spans $G_r'$ is taken:

$$score(P_r,G_r) = \max_{G_r'} \frac{2 \cdot |P_r \cap G_r'|}{|P_r| + |G_r'|}$$

\noindent The evaluation metrics most commonly reported are the total precision, recall, and F1 of this score function over all $R$ role instances:

\begin{align*}
precision &= \frac{\sum_r^R score(P_r,G_r)}{\sum_r^R \mathbb I[|\{P_r'\}| \geq 1]} \\
recall &= \frac{\sum_r^R score(P_r,G_r)}{\sum_r^R \mathbb I[|\{G_r'\}| \geq 1]} \\
\end{align*}

\noindent where $\mathbb I$ is an indicator function, $\{G_r'\}$ is the set of all gold spans for role instance $r$, and $\{P_r'\}$ is the set of all predicted spans for $r$ (usually a singleton set). The denominator of the precision thus counts only those role instances for which there is at least one predicted span and the denominator of the recall counts only those cases where there is at least one gold span. For simplicity and ease of comparison, I focus primarily on F1, reporting other metrics only where they are especially illuminating. The remainder of this subsection describes notable approaches that have been applied to SemEval, beginning with the winning system from the original competition.

\subsubsection{Chen et al. 2010}
The overall best performing system in the original competition was based on a discriminative, probabilistic frame-semantic parser called SEMAFOR \citep{chen2010semafor, das2010probabilistic, das2010semafor}, which performs ordinary SRL as a preliminary step and only then undertakes NI linking. For SRL, the system begins by using hand-designed rules to identify trigger tokens within the sentence that are likely to be FrameNet LUs. It then identifies the most likely frame for each trigger token using a logistic regression model based on lexico-semantic features of the token, features of the candidate frame, and features of the token's similarity to a set of targets from the training set, dubbed ``prototypes,'' that are known LUs of the candidate frame. Once a frame has been selected, arguments for each of the frame's FEs are selected using a separate logistic regression model over all individual tokens in the sentence, along with all spans associated with valid subtrees of a silver dependency parse for the sentence. Features for this model are primarily syntactic---such as candidate argument POS tag, and relative word order and dependency relation between the candidate and the trigger---but they include semantic ones as well (e.g. the candidate-trigger WordNet relation).

For NI linking, SEMAFOR considers only \textit{core roles} for a frame that were not filled by local arguments in the previous step, and divides the task conceptually into two stages: (1) determining the type of the NI (definite, indefinite, or \textit{masked}), and (2) if definite, locating the (non-local) referent.\footnote{FrameNet defines the notion of a \textit{Coreness set}, or \textit{CoreSet} for short, as a group of FEs, such that providing an argument for any one of them is sufficient to yield a felicitious and informationally complete sentence. The FEs \textsc{SOURCE}, \textsc{GOAL}, and \textsc{PATH} are one such CoreSet. When one of the FEs in a CoreSet is filled by their model, \citet{chen2010semafor} classify the others as \textit{masked}.} In practice, the two stages are merged into one, with \textit{indefinite} and \textit{masked} treated as special candidate fillers alongside other non-local spans. SEMAFOR considers nouns, pronouns, and noun phrases from the three sentences preceding the trigger as candidates, and uses another logistic regression model to select the filler.\footnote{Considering only nouns, pronouns, and noun phrases is a far too aggressive heuristic; as noted by the authors, it limits oracle recall to 20\% of all DNIs in the dataset.} The features for the NI linking model differ from those for the SRL model only slightly, replacing the dependency-based features with ones based on (1) whether the head of the candidate span appears as an exemplar for the role in FrameNet's exemplar annotations, and (2) the maximum syntactic distributional similarity between the head and any head word from an exemplar.

Frustratingly, for task 2, \citeauthor{chen2010semafor} only report and discuss precision, recall, and F1 on the first stage of their two-stage breakdown of NI linking---classifying each NI as DNI, INI, or masked. The results for the second stage are recoverable from a confusion matrix for the full model only (task 3), but they claim the confusion matrix for task 2 is similar. Although performance on the first stage is reasonable (50\% and 56\% F1 on the two chapters of the test set, respectively), results for the second stage are poor, yielding a mere 2\% F1. This is driven by an abysmally low recall of 1\%, though precision is quite high (57\%). The authors attribute these results to the ``inherent difficulty'' of the task, which---while fair---is hardly an illuminating diagnosis. 

\subsubsection{Laparra and Rigau 2012}
Since \citet{chen2010semafor}, there have been only a handful of successful attempts to improve upon the baseline established by SEMAFOR for task 2. One effective line of attack consists in using information about \textit{explicit} core roles and their fillers in order to make inferences about the distributions of arguments for implicit ones. Given the relative immaturity of NLP research on argument linking, this is a sensible thing to do. But its usefulness as a general methodology is surely limited: the kinds of roles that tend to be left implicit likely differ systemically from those that to be explicit, and one would accordingly would expect their arguments to differ as well.

Regardless, \citet{laparra2012exploiting} use a pipelined approach in this vein that leverages information about instantiated FEs in the training data to learn per-FE joint distributions $P(s,p)$ over the FrameNet semantic type ($s$) and the part of speech ($p$) of the head of the filler. It learns these distributions by simply counting co-occurrences in the training data. When processing a particular LU and its frame at evaluation time, the model identifies the most common \textit{pattern} of FEs associated with that frame in the training set that is satisfied by the LU's instantiated FEs.\footnote{A pattern is simply an ordered set of FEs. For instance, the \texttt{Residence} frame has a pattern \texttt{Resident Location} and a pattern \texttt{Resident Co\_Resident Location}, among others.} It then labels all uninstantiated core FEs as DNIs and resolves the DNIs by selecting the token that maximizes $P(s,p)$ over all tokens within the sentence containing the LU and the two preceding it. With this method, they considerably outperform SEMAFOR on task 2, obtaining an F1 of 19\%.

\subsubsection{Silberer and Frank 2012}
\citet{silberer2012casting} take the view that argument linking is more like (zero) anaphora resolution than like SRL, which has some precedent in the early work of Martha Palmer \citep{palmer-etal-1986-recovering-implicit}. In contrast to other efforts on the task, their focus is thus on linking whole \textit{entities} rather than particular mentions. Given a target DNI instance $d_k$ and some subset $\mathcal{E}_k \subseteq \mathcal{E}$ of relevant entities in the document (represented as a set of gold coreference chains), the goal is then to select a single entity $e_j \in \mathcal{E}_k$ as the filler for $d_k$. The most immediate question that arises in this setup is how to identify the set $\mathcal{E}_k$, and they consider three possibilities:
\begin{description}
    \item[AllChains] $\mathcal{E}_k$ is the set of all entities mentioned in the document up to the LU.
    \item[SentWin] $\mathcal{E}_k$ is the set of entities that have mentions with a particular syntactic type (NP, PP, ADVP, VP, S) occurring within the sentence containing the trigger and the two preceding it.
    \item[Chain+Win] $\mathcal{E}_k$ includes all the \textbf{SentWin} entities, plus any entities mentioned at least five times up to the LU.
\end{description}
Once $\mathcal{E}_k$ has been determined, they use a Bayes Net classifier to compute the probability of each mention $e_j \in \mathcal{E}_k$ being the correct referent of the DNI $d_k$. The features for the Bayes Net are mostly coreference-based features, but also include ones that incorporate SRL information. In the former group are features such as the minimum sentence and mention distances between the LU and any of an entity's mentions, as well as a \textit{prominence score} that scores the importance of an entity $e_j$ relative to others in the context window $w$:
\begin{align*}
    prom(e_j,w) &= |mentions(e_j,w)| - avg\_prom(w) \\
    avg\_prom(w) &= \frac{\sum_{e_{j'} \in \mathcal{E}_k |mentions(e_{j'},w)|}}{|\mathcal{E}_k|}
\end{align*}
In the latter group are features such as the concatenation of the frame and role names, and the semantic type of the DNI concatenated with the semantic types of the explicit roles.

Apart from their framing of argument linking as coreference, \citeauthor{silberer2012casting}'s other major contribution is their method for data augmentation. Using several existing datasets annotated for both entity coreference and SRL, they create new examples for task 2 by deleting anaphoric pronouns and reassigning their role label to the most recent coreferent mention preceding the LU, as shown in the following example:\footnote{The datasets include OntoNotes 3.0 \citep{hovy2006ontonotes}, ACE-2 \citep{ace2}, and MUC-6 \citep{muc6}. OntoNotes uses PropBank \citep{palmer2005proposition}, and so the FrameNet roles are obtained by mapping using SemLink \citep{palmer2009semlink}. ACE-2 and MUC-6 lack any SRL information, and so the authors apply SEMAFOR to both to generate silver SRL annotations.}

\begin{displayquote}
a. $\text{Riady}_k$ spoke in $\text{his}_k$ 21-story office building on the outskirts of Jakarta. [...] The timing of $\underline{\text{his}}_{k,Speaker}$ $\underline{\text{statement}}_{Statement}$ is important.

b. $\text{Riady}_k$ spoke in $\underline{\text{his}}_{k,Speaker}$ 21-story office building on the outskirts of Jakarta. [...] The timing of $\emptyset$ $\underline{\text{statement}}_{Statement}$ is important.
\end{displayquote}

This yields only crude, generally ungrammatical approximations of true cases of DNI, but it allows them to boost the size of their training data by well over an order of magnitude.

In their experiments, the authors evaluate in two settings. Since their focus is principally on resolving cases of DNI (and not on classifying types of NI), one of the settings is a strict subset of task 2, in which they have gold DNI annotations in addition to their gold coreference chains and SRL data. Since there were no existing systems that evaluated in this same setting, they compare against a heuristic baseline that always selects the entity with the highest prominence score. This model manages a test F1 of 20.5\%, while their best Bayes Net model---trained on the SemEval data augmented with OntoNotes only and using the ``Chains+Win'' method for determining candidate entities---achieves 27.7\%.

For more direct comparison with previous work, they also evaluate on the full task 2. But as their model only does DNI resolution, they have to outsource the initial NI classification component of task 2 to another model. For this, they use a support vector machine (SVM) with hand-selected features that include the FrameNet semantic type of the implicit role, the voice of the trigger (active or passive), and the relative frequency of DNIs vs.\ INIs for the role in the training set. Their best model in this setting is the same one as in the first, and it dramatically outperforms SEMAFOR, but falls just shy of \citep{laparra2012exploiting}, reaching 18.0\% test F1.

\subsubsection{Roth and Frank 2013}
\citet{roth2013automatically} combine elements of the two previous approaches, relying on coreference data while also leveraging information about explicit roles. They use a corpus of predicate pairs that they automatically align via graph clustering across pairs of documents (the \textit{R\&F} dataset; see \citet{roth2012aligning}) that concern the same events in order to recover arguments for roles that are implicit in one document but that are explicit in the other. Each predicate is a node in the graph and edges connect predicates across documents; edge weights are given by a linear combination of four predicate similarity scores, some of which take into account argument similarity as well. The resulting graph is bipartite and clustering is performed via recursive min-cut operations down to a minimum cluster size of two predicates. They then apply the Stanford Coreference System \citep{lee2013deterministic} on the concatenated pair of documents to obtain entity coreference chains and the MATE tool suite \citep{bohnet2010top} to obtain labeled semantic roles. With their aligned predicates and with this silver data, an argument may be linked to its implicit role(s) in the former document by tracing the coreference chain back from the explicit role in the latter. Though this method fails to outperform either \citet{silberer2012casting} or \citet{laparra2012exploiting} on task 2, it still bests \citet{chen2010semafor} with an F1 of 12\%. Despite these weaker results, the idea of aligning parallel text is intriguing, and deserves further investigation. I return to this idea in \S\ref{sec:future_work}.

\subsubsection{Feizabadi and Padó 2015}
Yet another approach to task 2 aims to directly tackle the paucity of training data via augmentation. Arguably, \citet{roth2013automatically} is an instance of this approach as well, but \citet{feizabadi2015combining} take the further step of literally combining existing corpora from different domains, as opposed to merely aligning them. To the SemEval dataset they add the entirety of the Beyond NomBank (BNB) training split (see \S\ref{subsec:nombank}) as training data, which requires some preprocessing, as SemEval uses the FrameNet role ontology where BNB uses the PropBank one \citep{palmer2005proposition}. To handle this, they map the SemEval annotations into PropBank using a script provided by the SemEval organizers, and then evaluate their system only against the PropBank ontology. Unfortunately, this makes direct performance comparisons to the previous systems impossible.

The model first searches for all instances of the target trigger in the OntoNotes dataset \citep{hovy2006ontonotes}, which provides PropBank annotations on a large and diverse text corpus. The model then extracts the most common set of roles for that trigger across all of its instances in OntoNotes. Linking is then performed on all roles in this set that are not already filled by local arguments, using a Naïve Bayes classifier with ten syntactic and semantic features to do binary classification independently for each $\langle$trigger, implicit role, candidate argument$\rangle$ triple. Structuring the model this way would permit multiple predicted arguments for each role, but it is unclear whether they take this approach or instead retain only the candidate with the highest posterior probability. Viable candidate arguments are all tokens and syntactic constituents from the sentence containing the trigger and the two preceding it.

With the earlier caveat that their results are not directly comparable to the previous efforts, the authors obtain an F1 of 19\% in this modified version of task 2. Interestingly, in an ablation that varies the amount of BNB data used during training, the benefits of their data augmentation appear to max out at just 10\% of the total size of the BNB training split. For comparison, using only SemEval training data without augmentation from BNB yields an F1 of 13\%. 

\subsection{Beyond NomBank}
\label{subsec:nombank}
\begin{figure}
    \centering
    \includegraphics[scale=0.5]{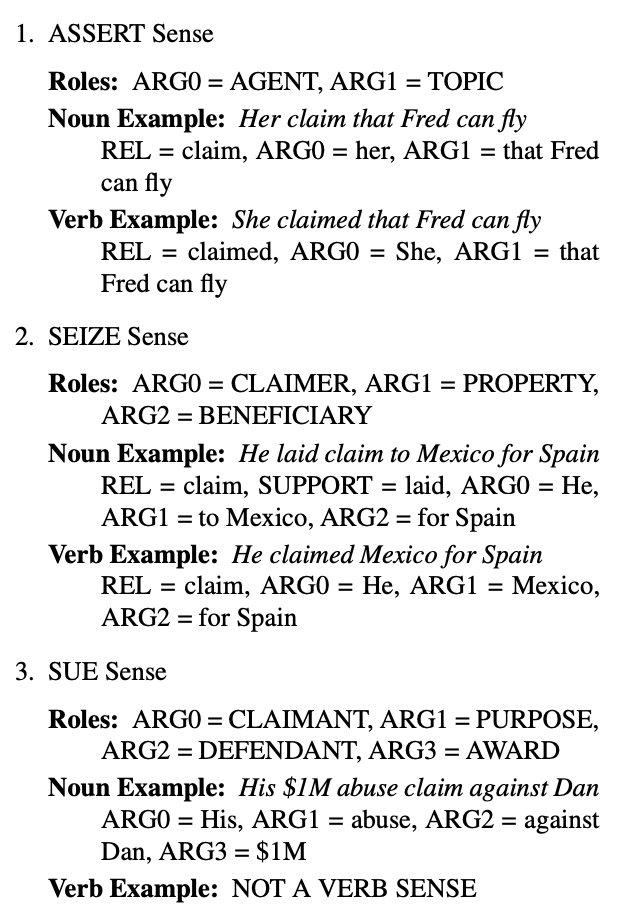}
    \caption{A NomBank frame for the noun \textit{claim}, with \textbf{Verb Examples} from PropBank overlaid.}
    \label{fig:nombank_frame}
\end{figure}

NomBank \citep{meyers2004nombank} is a sister project to PropBank \citep{palmer2005proposition} that annotates the local arguments of roughly 5000 common nominal predicates in the Penn Treebank II corpus. An example NomBank annotation is shown in Figure \ref{fig:nombank_anno}. In addition to direct annotations on Penn Treebank sentences, NomBank also provides lexical entries (frames) for each noun, just as PropBank does for verbs (see Figure \ref{fig:nombank_frame}). Both corpora use the Dowty-inspired numbered roles ARG0, ARG1, ..., ARG5, with ARG0 and ARG1 broadly corresponding to Dowty's proto-agent and proto-patient, respectively, and with the remaining ones corresponding to more predicate-specific roles.

\begin{figure}
    \centering
    \textit{Her gift of a book to John.} [NOM] \\
    \textsc{REL} = \textit{gift} \textsc{ARG0} = \textit{her} \textsc{ARG1} = \textit{a book} \textsc{ARG2} = \textit{to John} 
    \caption{A sample NomBank annotation, adapted from \citet{meyers2004nombank}. Note that the argument labeling scheme is the same as in PropBank.}
    \label{fig:nombank_anno}
\end{figure}

Motivated by NomBank's lack of \textit{implicit} role coverage, \citet{gerber2010beyond} annotate 10 predicates from NomBank for implicit roles to produce the Beyond NomBank (BNB) dataset. To select the predicates, the authors limited consideration to predicates with both verbal and nominal forms, and ranked these based on the product of two factors: (1) the average difference between the number of explicit roles in the verbal and nominal forms, and (2) the nominal form's frequency in the corpus. They use this score as an indicator of the likely prevalence of implicit arguments for a given predicate and select the top ten on the basis of that score. The motivation here is simply that common predicates for which roles often go unfilled in their nominal forms will yield the greatest amount of data for the task. Following is an example BNB annotation for the predicate \textit{investment}, in which \textit{all} of the predicate's roles are implicit:

\begin{displayquote}
\textins{$iarg_0$ Participants} will be able to transfer \textins{$iarg_1$ money} to \textins{$iarg_2$ other investment funds}. The \textins{p \textbf{investment}} choices are limited to \textins{$iarg_2$ a stock fund and a money market fund}.
\end{displayquote}

All arguments preceding the predicate in the document were considered viable candidates for annotation, and all instances of the ten selected predicates were annotated, resulting in 1,253 total instances with 1,172 cases of DNI. It is thus substantially larger than the SemEval dataset and has generally been the preferred benchmark for argument linking, mostly for this reason \citep{o2019bringing}.\footnote{BNB does not formalize the notion of INI as SemEval does. But one could very easily do DNI/INI classification on BNB, since the full set of roles is specified for all predicate instances, and one could just predict which ones are in fact filled.} Below, I present several notable argument linking models for BNB, beginning with \citeauthor{gerber2010beyond}'s own baseline model.

\subsubsection{Gerber and Chai 2010}
Like \citet{feizabadi2015combining}, \citeauthor{gerber2010beyond} treat argument linking as a binary classification problem: given a triple comprising the target predicate instance $p$, an implicit role $iarg_n$, and a gold coreference chain $c_{chain}$ for a candidate argument $c$, determine whether or not $c$ fills the role $iarg_n$ for $p$. For their model, they use a logistic regression classifier with 17 hand-engineered features. As with most of the SemEval models, these include syntactic features (e.g.\ the part of speech of $p$'s parent syntax node and the percentage of mentions $c' \in c_{chain}$ that are definite noun phrases), semantic features (e.g.\ the WordNet \citep{miller1998wordnet} synset of the head of every $c'$, concatenated with $p$ and $iarg_n$), as well as document-level features relating $p$ and $c$ or $c_{chain}$. The authors compare their model against a heuristic baseline that fills each $iarg_n$ for instance $p$ with the argument that fills $arg_n$ for some other instance $p'$ of the same predicate in the standard three-sentence context window. This heuristic essentially amounts to using only SRL information to do argument linking and it performs quite poorly: the baseline achieves 26.5\% F1 relative to the logistic regression model's 42.3\%.

It's noteworthy that even the results for these simple models are substantially better than any results on SemEval, and the reasons for this are over-determined. Not only does BNB have roughly twice the number of DNI examples, but it focuses on a much smaller collection of predicates, giving models more of an opportunity to learn the argument structures specific to those predicates. Moreover, BNB's use of PropBank-style roles---of which there are only six---means that models do not have to contend with nearly as much data sparsity as those evaluated against SemEval and its much larger set of FrameNet roles.

\subsubsection{Gerber and Chai 2012}
\citet{gerber2012semantic} later substantially improved their baseline model by expanding their feature set---from 17 to 81---grouping these features by type:
\begin{enumerate}
    \item \textit{Textual Semantics}: Features relating to the candidate filler $c$ and the role to be filled. Their most important feature by far, a concatenation of $p$, $iarg_n$, $p_f$, and $arg_f$, is one of this kind.
    \item \textit{Ontologies}: Features derived from specific ontologies, including FrameNet, VerbNet \citep{schuler2005verbnet}, and WordNet \citep{miller1998wordnet}. For instance, the \textit{frame relation path} in FrameNet between $arg_f$ and $arg_n$.\footnote{FrameNet specifies a number of different kinds of relations that can obtain between frames, such as parent/child inheritance, causality, and temporal precedence, among many others. SemLink \citep{palmer2009semlink} enables the mapping from PropBank to VerbNet and from VerbNet to FrameNet, from which the authors obtain these frame relations.}
    \item \textit{Filler-Independent}: Features that are unrelated to the coreference chain $c_{chain}$ and that usually depend only on $p$. These include a number of syntactic features of $p$ and its siblings in the constituency parse for the sentence containing it.
    \item \textit{Corpus Statistics}: These features are largely derived from the Gigaword Corpus and include statistics about corpus-wide frequencies of predicates and candidate arguments.
    \item \textit{Textual Discourse}: Features marking discourse relations, derived from the Penn Discourse TreeBank \citep{prasad2008penn}.
    \item \textit{Other}: Everything else. This category includes, for example, the sentence distance between the sentence containing the predicate and the one containing the candidate argument $c$, as well as the concatenation of $c$ with $p$ and $iarg_n$.
\end{enumerate}
In expanding their feature set, they found that those based on textual semantics (category 1)---and on SRL information, in particular---are most important, but that they are insufficient on their own (cf.\ the heuristic baseline from \citet{gerber2010beyond}). They also found that textual discourse features (category 5) contribute very little and can likely be dispensed with entirely.

The second notable change from their \citeyear{gerber2010beyond} work is their training and evaluation methodology. Rather than evaluate on a fixed test set, they perform a 10-fold cross validation, which enables them to use a larger number of examples during training. This, in conjunction with their expanded feature set, yields an improvement of 8.1\% F1 (to 50.3\% F1, macro-averaged across folds) over their original model. Moreover, they boost role coverage (the number of roles filled) by 71\% relative to the original NomBank annotations.

\subsubsection{Laparra and Rigau 2013}
Curiously, no one has been able to improve on \citeauthor{gerber2012semantic}'s cross-validated 50.3\% F1, which is suspicious considering the amount of effort others have put in to doing so, and considering that nearly 10 years have now passed since then. \citet{o2019bringing} went so far as to reimplement the heuristic baseline from \citet{gerber2010beyond} and obtained just 10.6\% test F1, contrasted with their reported 26.5\%. He contends that the most likely explanation for the discrepancy---and for the lack of improvement on the baseline more broadly---lies in potential differences in evaluation methodology, owing to the lack of standardized evaluation script and the ``sufficient ambiguity'' of the release format. Regardless, a number of more recent systems are still worth discussion simply for insight into the variety of methods that have been brought to bear on the task.

\citet{laparra2013impar} is the first of these. Continuing with their general program of leveraging information about arguments of explicit roles to fill implicit ones, they present a novel deterministic algorithm, ``ImPar,'' for argument linking on BNB. The core of their method consists in maintaining a \textit{default} argument for each of a predicate's roles that fills the role in the absence of a local argument. The default for a particular role is simply the most recent local argument that filled it. The authors justify this heuristic by an appeal to \textit{discourse coherence}, contending that, in general, recent instances of the same predicate will tend to refer to the same event and thus will be likely to have the same arguments---like the predicate \textit{losses} in the following passage:

\begin{displayquote}
\textins{$arg_0$ The network} had been expected to have \textins{$np$ \textbf{losses}} to \textins{$arg_1$ of as much as \$20 million} \textins{$arg_3$ on baseball this year}. It isn't clear how much those \textins{np \textbf{losses}} may widen because of the short Series.
\end{displayquote}

In many cases, however, there will be no antecedent local argument, and thus no default, to fill a particular role. In such cases, \citeauthor{laparra2013impar} select a candidate from the standard context window on the basis of a \textit{salience score}. First, they filter from the candidate set all local arguments of the predicate, all syntactic dependents, and all candidates that do not match the semantic category of the target role, where these categories are taken from the CoNLL 2008 dataset \citep{surdeanu2008conll} and map to sets of WordNet supersenses. Next, the salience score is calculated for all remaining candidates as a weighted sum of scores for six features: sentence recency ($w=100$) and five features concerning the syntax of the candidate, including whether it is a subject ($w=80$), a direct object ($w=50$), an indirect object ($w=40$), a non-adverbial ($w=50$), or the head of its constituent ($w=80$).\footnote{Both the features and their weights were taken from \citep{lappin1994algorithm}, which presented an influential algorithm for pronominal anaphora resolution.} The candidate with the highest salience score is then selected, and it furthermore serves as a kind of pseudo-default for subsequent instances of the role until either a local argument or a new (non-local) candidate with higher salience score is found. Empirically, the authors also find that applying a distance-based discount factor to the sentence recency component of the score for pseudo-default arguments helps to appropriately filter them from consideration for predicate instances occurring much later in the document.

Although ImPar fails to surpass the 81-feature logistic regression model of \citeauthor{gerber2012semantic}, it still outdoes their 17-feature one with a test F1 of 45.8, which is especially impressive considering the deterministic nature of the algorithm.

\subsubsection{Schenk and Chiarcos 2016}
Like \citeauthor{laparra2013impar}, \citet{schenk2016unsupervised} think that the data sparsity issue in argument linking can be best addressed by relying on local arguments. But unlike \citeauthor{laparra2013impar}, they propose to do so in an unsupervised manner. In particular, their aim is to learn predicate-specific role embeddings to represent the prototypical filler, or \textit{protofiller}, for that role by maximizing the cosine similarity between the protofiller embedding and the embeddings of all tokens in the training corpus belonging to some constituent that fills the role. Generating the protofiller embeddings just amounts to mean-pooling over all embeddings for such tokens. At inference time, selecting a filler is simply a matter of identifying the constituent within the context window whose (mean-pooled) embedding maximizes the cosine similarity with the protofiller. To narrow the set of candidates, constituents with syntactic categories that never fill the implicit role in the training set are excluded from consideration.

To obtain training data, they run existing frame-semantic parsing tools over two corpora, the \textit{Corpus of Late Modern English Texts} \citep[CLMET]{de2005corpus} and English Gigaword \citep{gigaword}, yielding silver sentence-level FrameNet annotations in the former case and BNB annotations in the latter, and gold constituency parses are available for both.

The authors evaluate on both BNB and SemEval and experiment with a variety of off-the-shelf embeddings. On BNB, their performance is uncompetitive, with F1 scores in the mid 30s. On SemEval, though, they do much better relative to other efforts, reaching 26.4\% test F1 and coming quite close to the state-of-the-art of 27.7\% held by \citet{silberer2012casting}. While the reliance on gold syntactic parses and on SRL tools for silver SRL data somewhat undermines \citeauthor{schenk2016unsupervised}'s claim that this is a purely unsupervised approach, it's nonetheless an attractive way to make use of large unlabeled corpora for argument linking, which is an idea I return to in \S\ref{sec:future_work}.

\subsubsection{Cheng and Erk 2018}
\begin{figure}
    \centering
    \begin{tabular}{c|c}
         \includegraphics[scale=0.65]{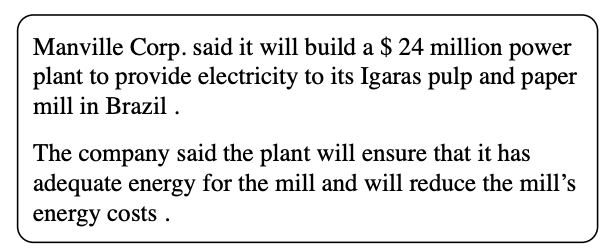} & \includegraphics[scale=0.65]{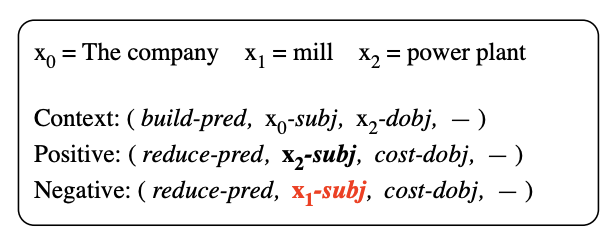}
    \end{tabular}
    \caption{A sample passage (left) and training example (right) from \citep{cheng2018implicitevent}.}
    \label{fig:cheng_erk}
\end{figure}

The work of \citet{cheng2018implicitevent} brings a couple of novelties to the BNB task---perhaps because they didn't set out to do argument linking in the first place. Rather, they are interested in \textit{syntactic} argument prediction as a cloze task, in which the training objective is to correctly restore a syntactic argument that has been removed from the text by selecting an appropriate entity. They formalize an entity as a coreference chain (similar to \citet{gerber2010beyond}) of length at least two, and an event as a four-tuple comprising a verbal predicate $v$, a subject $s$, a direct object $o$, and a prepositional object $p$, where some of the arguments may be null (see Figure \ref{fig:cheng_erk} for an example). At inference time, one of the (non-null) arguments ($s$, $o$, $p$) is randomly removed from an input event and the model must recover the appropriate entity. Cloze objectives of this sort are hardly new in NLP or IE, but their use as a means of generating synthetic data for argument linking certainly appears to be.

The second novelty consists in their fairly successful application of neural networks to the problem. To obtain enough data to train the network, \citeauthor{cheng2018implicitevent} generate examples from over five million documents from English Wikipedia. They divide the documents into paragraphs and run Stanford CoreNLP \citep{manning2014stanford} over each to obtain silver dependency parses and coreference chains, keeping all verbal predicates with corpus frequencies of 500 or greater. A single training example comprises three events, represented as four-tuples of the form described earlier: the ``target'' (``positive'') event $e_p$, one of whose arguments is to be predicted; a ``context'' event $e_c$ that is from the same paragraph as the target and that features the same entity as the argument to be predicted; and a ``negative'' event $e_n$ that is identical to the positive tuple except that the target argument has been replaced with a random entity from the document.

\begin{figure}
    \centering
    \includegraphics[scale=0.6]{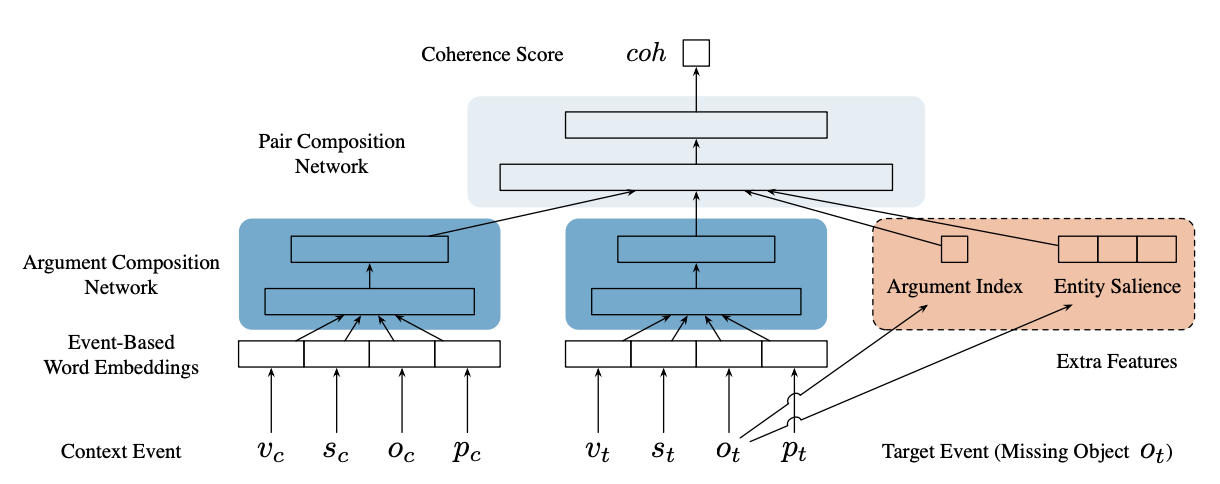}
    \caption{\citet{cheng2018implicitevent}'s \textit{event composition network} for computing a discourse coherence score between a context event and a target event. The illustration shows the target event's direct object ($o_t$) is the missing argument for this example.}
    \label{fig:cheng_erk_model}
\end{figure}

The training objective is similar in spirit to the negative sampling of \citet{mikolov2013advances}, but where the goal is to maximize the probability of event pairs that preserve \textit{narrative coherence}, rather than of word pairs that co-occur, while simultaneously diminishing the probability of event pairs that do not preserve coherence:

$$\frac{1}{m}\sum_{i=1}^m -\log(coh(e_{ci},e_{pi})) - \log(1 - coh(e_{ci},e_{ni}))$$

\noindent where $m$ denotes the number of training examples and $coh$ denotes a function of event pairs that outputs a \textit{coherence score}, and $e_c$, $e_p$, and $e_n$ are as defined above. The rationale is simply that the correct fillers for a particular role will tend to be those that best preserve the narrative coherence of the document. The function $coh$ is parameterized by a multi-component, feed-forward neural network, shown in Figure \ref{fig:cheng_erk_model}. To begin, the authors use  Word2Vec \citep{mikolov2013advances} to obtain ``event-based'' word-level embeddings for $v$, $s$, $o$, and $p$ in the event four-tuples. For a given sequence of events in a document, the predicate and argument tokens for all events in the sequence are concatenated and the Word2Vec algorithm is applied to the resulting text to produce the embeddings. The $v$, $s$, $o$, and $p$ embeddings are concatenated for the context event and for the target event, and the resulting vectors are separately passed through a feed-forward \textit{argument composition network} to produce embeddings for the two events. Independently, several other argument-based features are extracted, including the role type of the cloze argument, as well as a set of \textit{entity salience features} that aim to capture the relevance of the argument to the surrounding context. Finally, the two event embeddings and argument features are concatenated and used as input to a second feed-forward network, the \textit{pair composition network}, that outputs the coherence score $coh(e_c, e)$.

To fit this to an argument linking framework---which involves predicting the arguments for multiple \textit{implicit} roles at a time---the authors had to make a couple of modifications to their model. In particular, in the multiple implicit role setting of BNB, they must determine for each role whether it needs to be filled or not---essentially distinguishing cases of DNI from cases of INI. To do this, they use an auxiliary logistic regression classifier with a subset of the features used in \citep{gerber2012semantic}. To further aid the model in learning to associate particular entities with particular roles, they augment each training instance with additional negative events that have the same argument as the positive event, but that each assign that argument a different (incorrect) role. With these changes, \citeauthor{cheng2018implicitevent} achieve 49.6\% test set F1 and come very close to matching the performance of \citep{gerber2012semantic}. In fact, in some ways it is a more impressive result, both because \citeauthor{cheng2018implicitevent} are not leaning on the hand-engineered features used in the former work that depend on access to gold syntactic parses and ontology frames, and because their model was actually developed with a simpler, single-argument prediction task in mind.

\subsubsection{Cheng and Erk 2019}
Evidently unsatisfied with their earlier formulation of argument linking as a cloze task, \citeauthor{cheng2019implicitreading} reformulate the problem as reading comprehension in a follow-up paper (\citeyear{cheng2019implicitreading}). In one common reading comprehension setting, the goal is to produce the correct response to a query about a document by locating the relevant portion of text within that document. In recent years, reading comprehension models have commonly implemented some variation of the ``Attentive Reader'' architecture introduced by \citet{hermann2015teaching}. In broad strokes, this involves the following steps:
\begin{enumerate}
    \item Separately encoding all tokens in the document and all tokens in the query (typically with some form of RNN).
    \item Using the output vector representation for the query from (1) as the query vector in an attention mechanism over token vectors in the document to induce a distribution over those tokens.
    \item Selecting the token(s) in the document with highest attention score(s) as the response to the query.
\end{enumerate}

\begin{figure}
    \centering
    \includegraphics{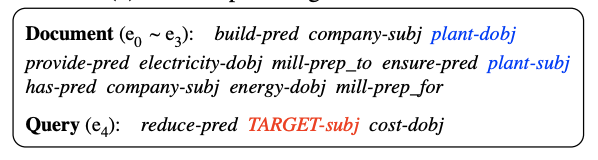}
    \caption{A sample query and document representation using the same passage as in Figure \ref{fig:cheng_erk}. Here, the target predicate is \textit{reduce} and the query requires the model to identify the subject (``the plant'').}
    \label{fig:cheng_erk_document}
\end{figure}

\begin{figure}
    \centering
    \includegraphics[scale=0.6]{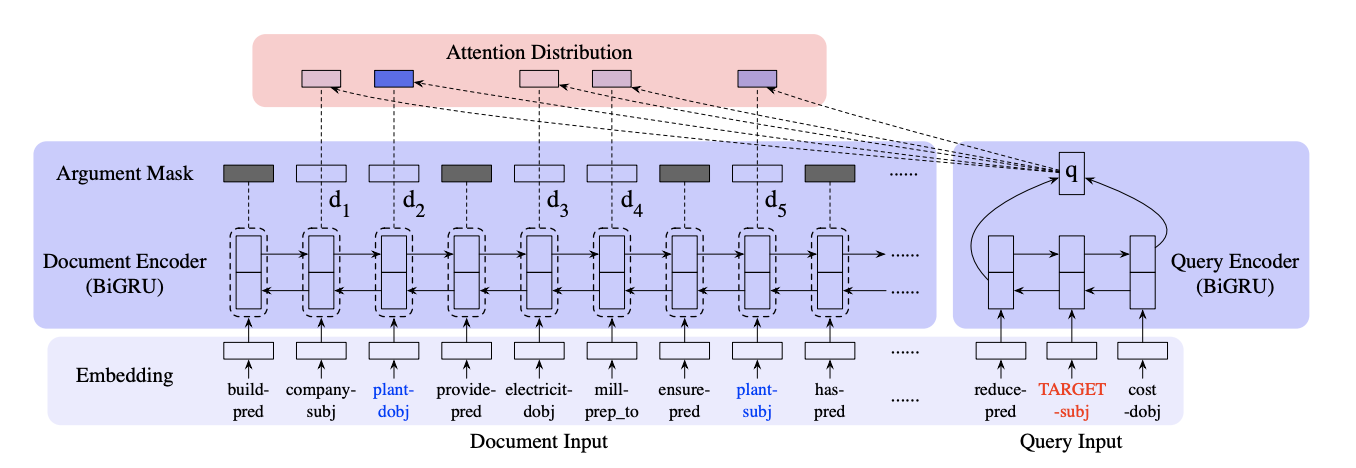}
    \caption{The Pointer Attentive Reader network from \citet{cheng2019implicitreading}, modified for argument linking from \citet{hermann2015teaching}. The attention distribution determines which of the arguments from the document context will be selected to fill the missing argument in the query. The further modification for pre-enriched query representations is not shown.}
    \label{fig:cheng_erk_attentive_reader}
\end{figure}

\citeauthor{cheng2019implicitreading} adapt this architecture as follows. For the input documents, they use the same concatenation of all predicates and arguments as they did in their original work. Similarly, for the query, they use a $(v,s,o,p)$ four-tuple with the target argument masked by a special \texttt{TARGET} token. An example input document and query (corresponding to the passage in Figure \ref{fig:cheng_erk}) are shown in Figure \ref{fig:cheng_erk_document}. Query and document word representations (for arguments only) are obtained from separate bidrectional GRUs \citep{cho2014learning}. For the query, the final hidden states from both directions are concatenated to form the representation $\bm q$, and for each document argument $\bm d_t$, the hidden states from the forward and backward directions from the corresponding time step ($t$) are concatenated. They then induce a probability distribution $\bm a$ over document argument tokens only using additive attention:

\begin{align*}
s_t &= \bm v^T \cdot \tanh(\bm W[\bm d_t; \bm q]) \\
a_t &= \text{softmax}(s_t)
\end{align*}

\noindent where $\bm v$ and $\bm W$ are learned. This network is shown in Figure \ref{fig:cheng_erk_attentive_reader}.

In practice, they find that the query representation---which has context only from the predicate and its other arguments---often doesn't contain enough information to be able to correctly identify the missing argument. This is especially the case in situations where other, non-target arguments are missing. To handle this, they use the attention computation above in conjunction with the contextualized argument representations from the document, to obtain a new query representation, pre-enriched with information about the relevant arguments in the document:

\begin{align*}
    \bm o &= \sum_{t=1}^T a_t \cdot \bm d_t \\
    \bm q' &= \bm o + \bm q
\end{align*}

\noindent The authors refer to this as ``multi-hop attention,'' as the attention-based query enrichment above could conceivably be repeated to encode more information for each missing but non-target argument, computing separate attention distributions over the document tokens for each one and obtaining new query representations $\bm q', \bm q''$, etc. In their implementation, however, they perform the above computation just once.

In many of the reading comprehension tasks to which similar neural architectures were applied, the correct answer to a query could be found only at a single location in the document. In contrast, with argument linking, any member of the appropriate coreference chain is arguably correct. To accommodate this difference, they maximize the log likelihood of the \textit{maximum} attention score across all coreferents of the correct argument. Using the same Wikipedia training data as in their \citeyear{cheng2018implicitevent} paper, \citeauthor{cheng2019implicitreading} obtain 48.3\% test set F1 on BNB---very comparable, though slightly inferior, results as compared to their first effort (49.6\% F1). This is perhaps unsurprising; although they have reframed the problem as reading comprehension, the objective is still fundamentally a cloze task and the training corpus is identical, even if we might expect some gain from the use of attention over candidate arguments. This could be some evidence of a problematic mismatch between the real objective---argument linking---and its operationalization as restoring masked \textit{explicit} arguments.

\subsection{ONV5}
\label{subsec:onv5}
\citet{moor2013predicate} present ONV5---a dataset very similar in spirit to BNB, but that targets a small set of verbal predicates rather than nominal ones.\footnote{The authors don't actually use this name for their dataset. Rather, it seems to be due to \citet{o2019bringing}, who claims it refers to the fact that the predicates are taken from the OntoNotes 5.0 corpus. But this is untrue; they come from the 4.0 corpus. The mystery remains, but I have stuck with \citeauthor{o2019bringing} for the sake of having a name to use.} Like \citeauthor{gerber2010beyond}, their primary concern is remedying the predicate-level annotation sparsity of SemEval, and thus privilege finding fewer predicates with high role recoverability over broad coverage of the lexicon. To construct the dataset, they begin with OntoNotes 4.0 corpus \citep{marcusontonotes}, which features ordinary PropBank-style SRL annotations, but they then map these into FrameNet using the script from the SemEval task and select predicates pairs based on the following criteria:
\begin{enumerate}
    \item No light verbs are permitted.
    \item Only predicates whose senses are covered by FrameNet, VerbNet, \textit{and} PropBank are permitted.
    \item Only predicates whose senses have a ``critical number'' of resolvable implicit roles are permitted.\footnote{Unrealized core roles in FrameNet are marked for definite vs. indefinite interpretations, which is what makes this criterion practicable.}
\end{enumerate}
Much about these criteria is nebulous. It is not at all clear, for instance, how light verbs were determined, what counts as a ``critical number'' of resolvable roles, or how the criteria collectively yielded a list of only five predicates: \textit{give} (with FrameNet frame \textsc{giving}), \textit{put} (\textsc{placing}), \textit{leave} (\textsc{departing}), \textit{bring} (\textsc{bringing}), and \textit{pay} (\textsc{commerce\_pay}). Annotation then consists in three tasks. First, for each predicate instance, determine whether its meaning in context actually matches that of the FrameNet frame. Then, for each instance that does, and for each NI of the instance, determine whether the NI is resolvable or not from context, and link resolvable NIs to their closest mention in the document.\footnote{There is also a middle, second step in the annotation that entails distinguishing \textit{lexically} licensed from \textit{constructionally} licensed NIs, but the authors give no details on this and it's irrelevant to the task.} In total, their protocol yielded 630 cases of NI across the five predicates, with just 242 resolvable.

With ONV5, they conduct two experiments. In the first, they use the same model and cross-validation setting as \citet{gerber2012semantic} to evaluate the model's linking ability when trained and tested on the same predicate. Here, they rely on gold constituency parses, and consider as candidates all constituents of type NPB, S, VP, SBAR, and SG within the context window, and they construct negative examples using local arguments. In this setting, they manage an average F1 of 36.4\% across the five predicates---significantly less than the models models on BNB.

In their second experiment, they use ONV5 as additional training data for the same DNI resolution-only subtask of SemEval task 2 as \citet{silberer2012casting}, and furthermore use their same model. Here, they obtain their best results by applying an unspecified feature selection procedure and are able to improve upon Silberer and Frank's performance, reaching 29.8\% test F1.

On the whole, ONV5 seems not to have made a lasting contribution to the argument linking literature. No one save \citet{o2019bringing} has directly evaluated on it, and even he was doing so in a cross-dataset probing context not aimed at setting a new state of the art. But the fact that ONV5 is natively compatible with FrameNet, VerbNet, and PropBank is not to be neglected, and its ability to serve as additional training data for argument linking using any of these ontologies is perhaps its greatest virtue. 

\subsection{Multi-Sentence Abstract Meaning Representation}
\label{subsec:ms-amr}
\begin{figure}
    \centering
    \includegraphics[scale=0.5]{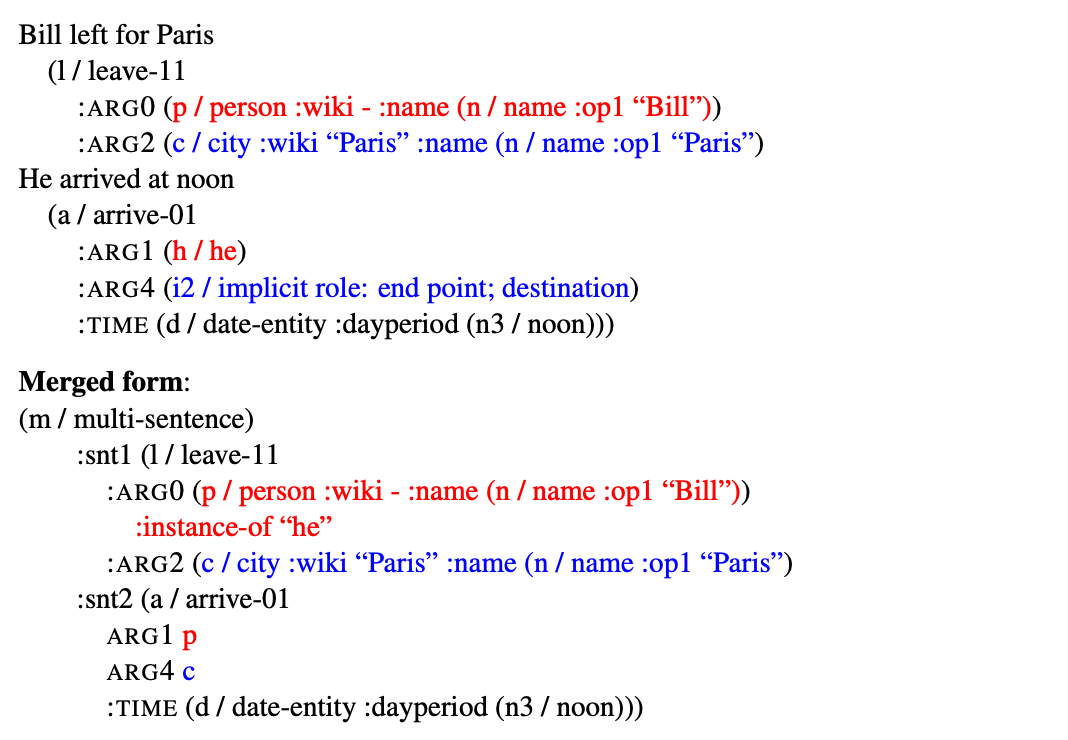}
    \caption{Figure from \citep{o2018amr} showing how traditional sentence-level AMR graphs are combined into document-level graphs in the MS-AMR corpus, along with annotations for coreferent events, entities, and implicit roles.}
    \label{fig:ms-amr}
\end{figure}

Abstract Meaning Representation (AMR) is a rich semantic formalism for representing the propositional content of sentences as rooted, directed graphs \citep{banarescu2013abstract}. Nodes correspond to entities in the sentence and edges describe the relationships between those entities. Among other things, the graphs capture sentence-level information about explicit semantic roles via PropBank-style numbered arguments but the corpus lacked similar information for implicit semantic roles until recently. To address this deficiency, \citet{o2018amr} introduce the Multi-Sentence AMR (MS-AMR) corpus, which provides a new annotation layer for cross-sentence coreference of entities and events on top of existing sentence-level AMRs. As the original AMR corpus annotates predicates with their PropBank senses, which include an associated list of roles, MS-AMR is furthermore able to provide annotations for coreferent implicit roles. Figure \ref{fig:ms-amr} shows an example. The top portion shows two sentence-level AMR graphs for the sentences \textit{Bill left for Paris. He arrived at noon.} The bottom portion shows the document-level graph, which merges the sentence-level ones, and adds annotations both for the entity coreference between \textit{Bill} and \textit{he}, and for the link between \textit{Paris} in the first sentence and the implicit destination role (ARG-4) of \textit{arrive} in the second.

The dataset is annotated on about 10\% of the AMR 2.0 release and comprises 293 documents, with an average 27.4 sentence-level AMRs per document. 

To date, the only work on argument linking for MS-AMR has been that of O'Gorman himself \citep{o2019bringing}. In his PhD thesis, he undertakes an impressively broad comparison of different types of models when trained and evaluated on the datasets discussed so far, MS-AMR included. In particular, he experiments with three very different model types. Model 1 (the ``interpretable model'') scores role-argument pairs via a linear combination of outputs from a feed-forward network, whose inputs are themselves the outputs of six simpler models. The simpler models are each targeted at a particular phenomenon related to argument linking, but are pre-trained on other, larger corpora---chiefly OntoNotes 5.0. These models merit a brief description.

\paragraph{Selectional Preference} In this context, the \textit{selectional preference} of an argument ``gives some measure of whether a particular entity would be likely to fill a particular role within an event'' \citep{o2019bringing}. There are many ways this might be operationalized, but \citeauthor{o2019bringing} implements a slightly modified version of an earlier neural model for selectional preference \citep{van2014neural}. The modified model is a feed-forward neural network that takes as input embeddings for the argument and for a $\langle$predicate, role$\rangle$ pair and outputs a selectional preference score. It minimizes a hinge-loss objective that involves a single positive $\langle$predicate, role, argument head$\rangle$ triple and a single negative one: $\max(0, margin + score(p,r,m_{+}) - score(p,r,m_{-}))$.

\paragraph{Narrative Schema} To that extent that a particular event fits into a larger story or \textit{narrative schema}, a schema-based model may help in filling implicit roles. \citeauthor{o2019bringing} views the Pointer Attentive Reader of \citet{cheng2019implicitreading} as the latest and best iteration of a long line of models for narrative cloze objectives. \citeauthor{o2019bringing} implements a much simpler, LSTM-based version of this model that does away with the attention mechanism.\footnote{To refer to this as a ``narrative schema'' model is somewhat misleading. There are no real schemas in play here---only a prior chain of event-role pairs.} A single-layer unidirectional LSTM encodes a sequence of event-role pairs, each represented with a single embedding that is learned during training. The last hidden state of the LSTM serves as the representation of the event sequence. The target implicit role (represented as another event-role pair) is encoded identically. Scores are computed as the dot product between the target role representation and the representation for a given event sequence. The model is trained primarily on the CoNLL-2011 shared task for coreference \citep[includes PropBank annotations]{pradhan2011conll} and uses a similar hinge loss to the selectional preference model, with one true event sequence for each implicit role and nine randomly sampled false ones.

\paragraph{Deictic Arguments}
\citeauthor{o2019bringing} notes that ``deictically present entites''--- particularly the speaker or addressee in a conversation---represent one very common type of filler for implicit roles. To capture these cases, he proposes a single feature that has value 0 for any non-deictic argument, and otherwise has a value equal to the sum of the selectional preference scores of all \textit{locuphoric} (first- or second-person) pronouns in its coreference chain. Clearly, this is not a model unto itself, but a rather an additional feature to be used as input to the aggregate downstream model.

\paragraph{Mention Referentiality}
Another indicator of the likelihood of an argument to fill an implicit role is its likelihood of being referred to \textit{at all}. Expletive or dummy pronouns, such as the \textit{it} in \textit{it's raining}, are classically \textit{categorically} non-referential, but there are many more cases where reference is simply unlikely and it's sensible to want to train a model to identify these cases.\footnote{\citeauthor{o2019bringing} cites \textit{the hat} in the noun phrase \textit{the man in the hat} as one example of an entity unlikely to be referred to again.} To model likelihood of referentiality, \citeauthor{o2019bringing} trains another feed-forward network to predict whether a given span should be considered a ``mention'' by predicting whether the span occurs in \textit{some} coreference chain. The network takes as input a set of traditional syntactic features for the mention (head word POS tag and lemma, dependency relation, and head word and dependency relation of each of the head's dependents) that are separately embedded and then averaged and passed to a final hidden layer that produces a scalar prediction.

\paragraph{Salience}
In contrast to the \textit{a priori} probability of an entity participating in a coreference chain, we might also consider how relevant it is to the current discourse context. This is what \citeauthor{o2019bringing} aims to capture with his salience model. He discusses a variety of ways in which the notion of ``salience'' could---and has---been cached out, but settles on the simple heuristic that the most salient entities are those most likely to be realized as pronouns. This is very similar in spirit to both the deixis and referentiality models, and also allows him to make further use of the OntoNotes corpus. The model itself is trained for simple coreference. It is yet another feed-forward network and accepts as input a vector of features of the target pronominal mention, its coreference chain, and a candidate mention. These features include a number of the same ones as the referentiality model (head word POS tag and lemma, dependency relation of the candidate) but also features for the length of the candidate's coreference chain and for grammatical parallelism with the target pronoun. All features are embedded and concatenated before being passed through the network, and binary cross-entropy is used as the loss.

\paragraph{Referential Status}
The final phenomenon \citeauthor{o2019bringing} considers is what he calls the ``referential status'' of an implicit role. This is really a composite of three interrelated aspects of an implicit role and its argument: (1) the kind of NI the implicit role represents (DNI, INI, CNI); (2) whether its argument is recoverable in the document context; and (3) the fine-grained type of the implicit role.\footnote{\citeauthor{o2019bringing} presents a taxonomy of 11 implicit role types, divisible into three coarser-grained categories that basically correspond to a clear DNI type, a clear INI type, and a ``somewhere in between'' type. See the thesis itself for details.} The model he proposes for referential status is accordingly a multi-task one for all three components. A shared feed-forward network is used across all three tasks, with individual classification heads for each one. The features used can be divided into those that characterize the implicit role itself and those that describe its syntactic context, all of which are categorical, and which are embedded before being passed through the feed-forward network.

\paragraph{}
\citeauthor{o2019bringing} performs many interesting intrinsic evaluations and ablations on the above models, but I refer the reader to his thesis for discussion of those. As mentioned above, Model 1 simply applies a final feed-forward layer to the outputs of some combination of the simpler models, sums \textit{those} outputs, and scales the result by a weight that is learned per implicit role \textit{construction type}. These types, proposed by \citeauthor{o2019bringing}, capture possible combinations of role type and syntactic context (e.g. ``nominal patient/experiencer''). If we suppose $M$ simple models are used and view each simple model as a function $f_k$ of a candidate argument $a_i$, a role $r_j$, and possibly some additional features $\phi_{i,j}$, Model 1 can be expressed as follows:

$$s(a_i, r_j, \phi) = \sum_{k=1}^M w_{construction(r_j)}\text{FFNN}(f_k(a_i,r_j,\phi_{i,j}))$$

\noindent where $w_{construction(r_j)}$ is the weight for the input role's construction type. \citeauthor{o2019bringing} dubs Model 1 the ``interpretable'' model, and that is only in virtue of these weights, as their final values can potentially give some indication of the relative importance of the constituent simple models.

Models 2 and 3 are conceptually less complex. Model 2 (the ``dense model'') is directly in the spirit of the early argument linking models and uses more traditional syntactic and semantic features (e.g. candidate POS tag, candidate head word) as inputs to a single feed-forward network. Model 3 (the ``ELMo model'') is an ELMo-based neural attention model \citep{peters-etal-2018-deep}. It begins by computing a preliminary score between roles and candidate arguments, using ``lightweight'' role and argument representations, $\mathbf{g}_r(j)$ and $\mathbf{g}_m(i)$, generated from the component feature models of Model 1, along with a separate vector $\phi_{i,j}$ for features of the relation between the two (e.g. sentence distance):
$$s_{preliminary}(i,j) = FFNN(\mathbf{g}_r(j)) \cdot FFNN(\mathbf{g}_m(i)) + FFNN(\phi_{i,j})$$

\noindent Where \textit{FFNN} here denotes a feed-forward network and ``$\cdot$'' denotes dot product. Based on this score, only the $k$ best candidate arguments are retained. A new, contextualized span representation $\mathbf{h}_m(i)$ is then generated for each of the $k$ best arguments by obtaining token-level embeddings for the argument's entire sentence using ELMo, applying a one-layer LSTM on top of these embeddings, and then taking an attention-weighted sum over the LSTM outputs for the tokens within the span. The same technique is used to generate a representation $\mathbf{h}_r(j)$ for the implicit role as well (by encoding the predicate with a feature for role number) and these are concatenated with the ``lightweight'' representations and used to compute the final score:

$$s_{final}(i,j) = s_{preliminary} + FFNN([\mathbf{h}_r(j); \mathbf{g}_r(j)]) \cdot FFNN([\mathbf{h}_m(i); \mathbf{g}_m(j)])$$

The argument with the highest final score is then selected as the filler. For each of the three models, \citeauthor{o2019bringing} considers various combinations of training corpus and test corpus for the corpora discussed so far. He trains individually on MS-AMR and BNB, but also jointly on all corpora (SemEval, BNB, MS-AMR, and ONV5).\footnote{\citeauthor{o2019bringing} is parsimonious on the details of how precisely the joint training is implemented.} In the all corpora setting, he also experiments with fine-tuning on the target corpus and separately with including sentence-level AMRs as training data. He evaluates in two settings: In the first, he both predicts which roles are recoverable and then links arguments for the recoverable roles; in the second, he assumes the recoverable roles are known and only links the arguments.\footnote{These settings map on to the two-stage breakdown of task 2 for SemEval that (e.g.) \citet{das2010semafor} use. The first setting involves doing both stages and the second setting only the second stage.} In both settings, he tests on SemEval, BNB, and MS-AMR, and additionally on ONV5 in the first setting. As MS-AMR is the focus of this section, I describe only those results, and they are poor across the board in both settings.\footnote{The results for the other datasets are generally uncompetitive with previous efforts, but may be of interest in evaluating the potential for transfer learning.} On both the full task and the linking only task, a version of Model 2 trained jointly on all corpora with fine-tuning on each achieves the best performance---a mere 8.25\% F1 in the former case and 29.03\% F1 in the latter. Model 3 trained and fine-tuned in the same manner approaches Model 2's performance on the full task, but no versions of either Model 3 or Model 1 come close in the linking-only task.
\begin{table}[]
    \centering
    \begin{tabular}{c|ccc}
    \toprule
     Corpus & candidates & recoverable & rate of recoverability \\
    \midrule
     MS-AMR & 30211 & 2206 & 6.8\% \\
     BNB & 3014 & 1028 & 34.11\% \\
     SemEval & 1591 & 298 & 18.7\% \\
     ONV5 & 390 & 213 & 35.3\% \\
    \bottomrule
    \end{tabular}
    \caption{Corpus-wide rates of recoverability of implicit roles according to \citet{o2019bringing}.}
    \label{tab:recoverability}
\end{table}

It's humbling that performance remains this low even with a powerful neural model like ELMo, but \citeauthor{o2019bringing} lays the blame on both the small size of the dataset and particularly on the fact that only 6.8\% of its implicit roles are recoverable. This is surely one of the main culprits. But it absolute terms, MS-AMR still has more than twice the number of recoverable roles than any of the corpora discussed so far (see Table \ref{tab:recoverability}). 

\subsection{Roles Across Multiple Sentences}
\label{subsec:rams}
\subsubsection{Ebner et al. 2019}
\citet{ebner-etal-2020-multi} introduce Roles Across Multiple Sentences (RAMS)---the largest dataset for argument linking to date, comprising 9,124 annotated events and covering nearly 4,000 documents. The annotations are based on an ontology developed as part of the DARPA AIDA program and focus on events related to recent conflicts between Russia and Ukraine. Examples in the dataset consist of a trigger span that evokes a typed event, and zero or more argument spans, each filling a role of the trigger. The annotation protocol considers arguments anywhere within a context window of five sentences---the sentence containing the trigger and the two on either side. This restriction was enforced to make annotation tractable and receives some further justification from \citeauthor{gerber2010beyond}'s observation that 90\% of implicit role arguments can be found within a sentence distance of two from the trigger \citep{gerber2010beyond}.

The authors additionally present a strong baseline model for RAMS, largely inspired by recent neural models for SRL \citep{he-etal-2018-jointly, ouchi-etal-2018-span}. They formalize the RAMS task as follows. Given a document $D$, an event $e$, and a set of roles $R_e$ that $e$ evokes, identify all role-argument pairs $(r,a)$ such that $r \in R_e$ and $a \in D$. This is the most general possible formulation of argument linking: it presupposes neither that each role must be filled by only one argument, nor that a single argument can fill only one role (see \S\ref{subsec:terminology}).\footnote{In this way, it's similar to the setup of binary classification over $\langle$trigger, implicit role, candidate argument$\rangle$ triples used by \citet{gerber2010beyond, gerber2012semantic} and \citet{feizabadi2015combining}.} Their modeling approach consists in learning contextualized span representations for the event trigger, roles, and candidate arguments, pruning the candidates, and then scoring each one relative to a particular event-role pair to select the argument for that role. Let $l(a,a_{e,r})$ denote the score for an argument $a$ relative to an event-role pair $\tilde{a}_{e,r} = (e,r)$ and let $A_e$ denote a set of candidate arguments for $e$, including the special null argument $\epsilon$. Then the model's learning objective is to maximize the probability of the true arguments for each event-role pair, where the probability is determined by a softmax, parameterized by the score $l$: 

$$P(a|e,r) = \frac{\text{exp}\{l(a,\tilde{a}_{e,r})\}}{\sum_{a' \in A_e}\text{exp }\{l(a',\tilde{a}_{e,r})\}}$$

\noindent In their best model, $l$ is actually a sum of two other scoring functions: $s_l(a, \tilde{a}_{e,r})$, given in (1) below, which directly scores a candidate argument relative to an event-role pair; and $s_{A,R}(a,r)$, given in (2), which scores a candidate argument relative to a role only. Both are parameterized as feed-forward networks ($\mathbf{F}$) with a final linear layer on top ($\mathbf{w}$):

\begin{align}
    s_l(a, \tilde{a}_{e,r}) &= \mathbf{w}^T_l\mathbf{F}_l([\mathbf{a};\mathbf{\tilde{a}}_{e,r};\mathbf{a} \circ \mathbf{\tilde{a}}_{e,r}; \phi_l(a,\tilde{a}_{e,r})])\\
    s_{A,R}(a,r) &= \mathbf{w}^T_{A,R}\mathbf{F}_{A,R}([\mathbf{a};\mathbf{r}])
\end{align}

\noindent where $\mathbf{a}$, $\mathbf{r}$, $\mathbf{\tilde{a}}_{e,r}$ are the contextualized span representations for the candidate argument, the role, and the event-role pair, respectively; ``$[\cdot;\cdot]$'' denotes vector concatenation; ``$\circ$'' denotes element-wise vector product; and $\phi_l$ is a hand-engineered feature vector. The authors liken $s_l$ to \citeauthor{schenk2016unsupervised}'s use of the cosine similarity between the embedding for a candidate argument and the embedding for the ``prototypical filler'' of the target role. In contrast, $s_{A,R}$ simply scores the candidate against a representation of the role itself, abstracted from any (prototypical) realization. The authors additionally experiment with similar score functions of both the event and the role ($s_{E,R}(e,r)$) and of the candidate argument and event ($s_c(e,a)$), but found in ablations that neither conferred any benefit.

Under the assumption that each role binds a single argument, the inference problem is to find the argmax of $P(a|e,r)$ independently for each event-role pair. But two additional decoding strategies are also considered: (1) \textit{greedy decoding} to identify the (potentially multiple) best argument(s) for a role, using $P(\epsilon|a,r)$ as a threshold; and (2) \textit{type-constrained decoding}, which can be used when gold event types and roles ($R_e$) are known, in order to select the top $k$ arguments for each role. Unsurprisingly, the authors achieve the best performance (75.1\% dev F1) using strategy (2), as they are able to leverage information from the ontology to narrow the search space.

\subsubsection{Chen et al. 2020}
Since the initial release of RAMS, there have been two attempts to improve \citeauthor{ebner-etal-2020-multi}'s baseline model. The first of these, proposed by \citet{chen2020joint}, formulates argument-linking in much the same way, but differs substantially in its modeling. Rather than treat arguments independently, \citeauthor{chen2020joint} \textit{jointly} model all candidate arguments together with their trigger using a Transformer-based network \citep{vaswani2017attention}. First, contextualized representations for an event trigger and all candidate arguments are obtained by concatenating pre-trained BERT embeddings \citep{devlin2019bert} for the start and end tokens in the span, along with a pooled attention vector over all tokens in the span, and applying a feed-forward network to the result. The resulting trigger representation $\bm t$ and candidate argument span representations $\bm m_1, \ldots, \bm m_n$ are then fed as a sequence $(\bm t, \bm m_1, \ldots, \bm m_n)$ to a Transformer encoder, with the traditional positional embeddings omitted. Finally, a linear layer atop the encoder produces for each argument a distribution $P(r|t,m)$ over all possible roles, where any roles that are invalid for the input event trigger are masked before normalization.

Two aspects of this model deserve careful notice. For one, it is only because the self-attention mechanism of the encoder is applied over all candidate arguments that this is a \textit{joint} model; there is no auxiliary training objective, for example, that enforces constraints on the number of arguments that may fill a given role or the number of roles a given argument may fill. For another, the distribution induced by the network, $P(r|t,m)$, is over \textit{roles}, given a particular trigger and argument---rather than over \textit{arguments}, given a particular trigger and role as in \citep{ebner-etal-2020-multi}. In principle, there is reason to prefer learning the former distribution, as the number of roles is virtually always fixed, whereas the number of candidate arguments grows quadratically in sentence length and roughly linearly in the size of the context window. However, the fixed-size context window of RAMS somewhat mitigates this problem.

With this model, \citeauthor{chen2020joint} are able to substantially outperform the original RAMS baseline, achieving a micro-averaged F1 of 79.9\% across roles, relative to \citeauthor{ebner-etal-2020-multi}'s 73.3\%. The model also demonstrates robust performance across sentence boundaries, as F1 for arguments in sentences outside the trigger sentence never dips below 74.4\%.

\subsubsection{Zhang et al. 2020}
Both of the aforementioned argument-linking models are span-based. In settings where gold argument spans are available, as in RAMS, this is not a problem. But in the more general and more common case, where span boundaries are unknown, one runs up against the quadratic number of candidate arguments discussed above. This concern has led \citeauthor{o2019bringing} to speculate that span selection models may simply not be feasible in such scenarios. Motivated by the same concern, \citeauthor{zhang2020two} introduce a two-step, Transformer-based model that ostensibly reduces the quadratic number of candidate spans to linear, and without pruning. The model first attempts to detect only the \textit{head words} of candidate arguments. For a given RAMS event trigger, contextualized representations of every word in the context window are obtained by feeding them through a BERT encoder, along with corresponding \textit{token type IDs} that differentiate (i) the event trigger word; (ii) other words in the trigger sentence; and (iii) words in non-trigger sentences. Using role-specific biaffine attention layers \citep{dozat2016deep}, it then determines a distribution over candidate arguments for a particular trigger and role by taking a softmax over the outputs of the biaffine layer for that role, when using as input the contextualized representations of the trigger paired with each word in the context window:

$$P_r(p,c) = \frac{\text{exp}\{\text{Biaffine}_{r}(\bm e_p, \bm e_c)\}}{\sum_{c' \in C}\text{exp}\{\text{Biaffine}_{r}(\bm e_p, \bm e_{c'})\}}$$

\noindent where the $C$ represents the union of all words in the context window, along with the null argument $\epsilon$. In the second step, the model determines the appropriate span for each argument by separately considering possible left and right span boundaries up to a distance $K$ away from the head. For both boundaries, the head and candidate boundary word representations, $\bm e_h$ and $\bm e_b$, are concatenated and used as input to a multi-layer perceptron (MLP), and a softmax is applied over the MLP outputs for each candidate to induce a distribution over possible left and right boundaries. The expression for the left boundary distribution is shown below, and the one for the right is analogous:

$$P_r(h,b) = \frac{\text{exp}\{\text{MLP}_{left}(\bm e_h, \bm e_b)\}}{\sum_{b' \in (h-K,h]}\text{exp}\{\text{MLP}_{left}(\bm e_h, \bm e_{b'})\}}$$

Cross-entropy loss is used for training during both steps, and the argmax decoding is used at inference time. For the purposes of direct comparison with \citet{ebner-etal-2020-multi}, the authors evaluate the model using gold spans. Since head words are not annotated in the RAMS corpus, the authors heuristically select as the head the word in the gold span with smallest arc distance from the root in a silver dependency parse for the surrounding sentence. In this setting, they are able to outperform the original span-based model without type-constrained decoding (71.0\% dev F1 vs. 69.9\%) and attain comparable performance with it (74.3\% dev F1 vs. 75.1\%). These results suggest that, at least for the purposes of this task, linking by head word alone may be sufficient while also drastically reducing the number of candidate arguments.

\section{Future Directions}
\label{sec:future_work}
This section aims to highlight some of the major limitations of the work described in $\S\ref{sec:arg_linking}$ and to speculate on possible ways beyond them. \S\ref{subsec:data_problems} focuses on data sparsity and possible approaches to augmentation and domain adaptation and \S\ref{subsec:model_eval_problems} focuses on issues in model design and evaluation.

\subsection{Data Sparsity and Augmentation}
\label{subsec:data_problems}
The chief lament of those who have worked on argument linking is unquestionably data sparsity---whether in the number of annotations per predicate or per role (SemEval), the number of recoverable roles (MS-AMR), or simply the number of training instances (essentially all corpora). There is every reason to think that this has been \textit{the} main bottleneck in making more rapid progress.

RAMS represents a genuine step forward in providing fully an order of magnitude more examples than any previous dataset. But it also has the distinct disadvantage of being tied to a highly domain-specific corpus and ontology, and we cannot be at all confident that models designed for such a specific domain will generalize well to others. Indeed, in her 2020 EMNLP keynote address, Claire Cardie, citing Dan Roth, observes that the performance of current systems for named entity recognition---surely a simpler IE task---drops 20-30\% when altering the domain, the genre, or the language \citep{cardie2020emnlp}. The domain adaptation experiments conducted by \citet{o2019bringing} corroborate this trend for argument linking.\footnote{These results are not reprinted here; see Table 5.9 in \citep{o2019bringing}.} Such complaints about insufficient data or model brittleness within machine learning and NLP are now cliché to the point of parody, but they're no less legitimate for that. So what can be done?

\subsubsection{Crowdsourcing}
Data sparsity in argument linking is precisely the focus of \citeauthor{feizabadi2019strategies}'s 2019 PhD thesis, and she advances two proposals. The first is to simply crowdsource the annotation of implicit roles. Crowdsourcing has the obvious benefit of rapidly accelerating the annotation process, but it requires extremely careful experimental design to do well, even for relatively straightforward annotation projects like sentiment analysis, named entity recognition, or question answering. This holds \textit{a fortiori} for (implicit) semantic role labeling, where one cannot just serve the task neat to untrained annotators, as one can essentially do in these other domains. Several significant task-specific challenges immediately present themselves:
\begin{enumerate}
    \item Translating the roles from a given ontology into an annotation format that is intelligible to a non-expert (e.g. natural language questions).
    \item Finding ways not to overwhelm or confuse annotators with too much text without compromising on recall of arguments.
    \item Reconciling different answer spans.
\end{enumerate}
These of course presuppose that other difficult questions about corpus and ontology selection and word sense disambiguation have been resolved.

Feizabadi's own forays into crowdsourcing argument linking annotation are illuminating on all of these fronts. Even to get the project off the ground, she found it necessary to radically restrict the problem domain by focusing only on predicates of motion. For her corpus, she uses Jules Verne's novel \textit{Around the World in Eighty Days}, both for its narrative structure and for its abundance of motion predicates. She extracts motion predicates using a simple predominant sense heuristic to perform word sense disambiguation, and then applies a combination of WordNet and FrameNet to identify triggers of motion-related frames, defined as any that share the same set of core roles as the \textsc{motion} frame itself, except for the \textsc{distance} role. Focusing only on these core motion-related roles enables her to manually map them to questions in a first experiment on Amazon Mechanical Turk (AMT).\footnote{Restricting the domain in this way is one way around the first of the three challenges I mentioned, but it clearly comes at the cost of generality.} Each human intelligence task (HIT) features a single trigger, highlighted in bold, and presented to users embedded in a three-sentence context window. Above the text are four questions corresponding to the four core roles of the motion frame: \textit{Where does the event take place} (\textsc{place}); \textit{What is the starting point} (\textsc{source}); \textit{What is the ending point} (\textsc{goal}); and \textit{Which path is used?} (\textsc{path}). Answers are given by entering text free-form into a box next to each question, with an optional checkbox to indicate that no answer could be found.

With five annotators per HIT, Feizabadi obtains just 37.7\% inter-annotator agreement (IAA) across roles using exact match, and 40.0\% when using word-based overlap. These results lead her to abandon this experimental setup, and her diagnosis of the low agreement is revealing:

\begin{displayquote}
The low result even in the overlap condition indicates that the problems can not [sic] be due mainly to minor
differences in the marked spans. Thus, we performed an analysis and realized that the main reason was that annotators were often confused by the presence of multiple predicates in the paragraph such that many marked roles pertaining not to the boldfaced target predicate but to other predicates. (76)
\end{displayquote}

\noindent This observation speaks directly to my second challenge and suggests that the trigger itself must be incorporated in the questions in order to stand a chance of obtaining high quality responses. It is also chastening to realize that even the usual context window of three sentences may be enough to distract annotators from the predicate under consideration.

To better focus the annotators, Feizabadi moves to a slot-filling setup in her second experiment. Retaining the same presentation for the trigger and context, she replaces the four questions from the previous setup with a template centered on the trigger:

\begin{displayquote}
Phileas Fogg \textbf{reached} from \_\_\_\_\_\_\_\_\_ (to) \_\_\_\_\_\_\_\_\_ through \_\_\_\_\_\_\_\_\_ path. The whole event is (was) taking place in (at) \_\_\_\_\_\_\_\_\_.
\end{displayquote}

\noindent where annotators are permitted to leave blank any slot for which they cannot find an argument. Even if less natural than the original experiment, this yields substantially improved agreement: 44\% across roles for exact match and 50\% for overlap. A further cause for optimism is that agreement levels are consistent when broken down by local, non-local, and unrealized roles. But the prevailing tone in Feizabadi's ruminations on the prospects for scaling her experiment is pessimistic. She suggests that the effort required to construct enough templates to achieve even modest coverage of (sets of) FrameNet frames is too high to be practicable. The prima facie tempting alternative of using the simpler role ontology of PropBank poses its own set of challenges on further reflection---most notably, the predicate-specificity of the interpretations that those roles receive. Moreover, agreement at the granularity of individual roles appears highly variable: average exact match agreement with the canonical annotation for \textsc{path} was an impressive 82\%, while \textsc{place} was only 50\%. Such variability would only grow if a broader set of roles were considered.

I share Feizabadi's doubts about the feasability of large-scale crowdsourced annotation for argument linking, and I do not have ready solutions for the issues she raises. That is hardly to say that it cannot be done, only that it will clearly demand significant ingenuity and engineering effort on the part of the researchers. But if that effort is considerable, it may turn out to be best to just invest it directly in annotation and secure the further quality benefits of expertise.

\subsubsection{Combining corpora}
Ultimately, Feizabadi finds more hope in her second proposed approach, which is domain adaptation via combination of existing corpora, as she and Padó attempted with SemEval and BNB in their \citeyear{feizabadi2015combining} paper. The results presented in the thesis are largely the same as those presented in that paper, and I don't repeat them here. However, it's worth considering in more detail the major difficulty of this approach that Feizabadi spotlights: reconciling different ontologies.

In certain cases, as here, there are tools to mitigate the problem, such as the PropBank conversion script provided by the SemEval organizers and SemLink for mapping between VerbNet and PropBank or between VerbNet and FrameNet. But there are also deeper problems relating to differences in the corpora themselves. BNB annotates only for verbal predicates, while SemEval includes nominal, adjectival, and adverbial ones as well. BNB annotates newsire and financial documents while SemEval annotates \textit{Sherlock Holmes} stories. From one angle, such corpus diversity is precisely what one wants in domain adaptation. But from another, there are likely significant limits to how useful one corpus will be in predicting the other, given their limited size. Indeed, \citet{feizabadi2015combining} do no better on either SemEval (19\% test F1) or on BNB (21\% test F1) than other systems that do not use combined corpora.\footnote{As discussed earlier, neither these results nor \citeauthor{o2019bringing}'s (2019) are directly comparable to other results on SemEval, as both authors convert SemEval's FrameNet annotations to PropBank. But an indirect comparison is still informative.} And yet the versions of their models that are trained exclusively on the training set for the target corpus do worse than those that use a mixture of both. By themselves, these results just seem too ambiguous to settle the question of the utility of this kind of domain adaptation. The only other work to have combined corpora for training is \citet{o2019bringing}, and his results do tilt the evidence somewhat in Feizabadi's favor. \citeauthor{o2019bringing} obtains his best result of 18\% F1 on SemEval when training Model 1 on BNB and his best of 31\% F1 on BNB when training Model 3 on all corpora (SemEval, BNB, ONV5, and MS-AMR) and fine-tuning on BNB.

\subsubsection{Alternatives}
There are other, often easier ways to do data augmentation. The most common flavor of augmentation among the works surveyed in \S\ref{sec:arg_linking} consisted in using a combination of SRL and coreference data. \citet{silberer2012casting} presented one promising and extremely simple method in this vein that entails reassigning roles of local arguments to the previous mention in the argument's coreference chain. This enabled them to boost the size of their training data by a factor of 10 and nearly match the state of the art set by \citet{laparra2012exploiting}. One may wonder why 10x more data does not push them \textit{beyond} the state of the art, but we should remember that they used silver SRL and coreference data obtained from systems that have since been radically outperformed at both tasks. Revisiting this method with current top-performing models seems likely to yield significant gains. Moreover, for coreference chains that extend beyond a length of two, their technique might be generalized to relabel mentions beyond the one immediately preceding the local argument.

Another tactic along these lines was \citeauthor{roth2013automatically}'s (\citeyear{roth2013automatically}) use of predicate-argument structures that were automatically aligned across documents on the same topic. The idea here was to use silver SRL and coreference annotations to link arguments of a predicate's explicit roles in one document to an aligned predicate's implicit roles in the other. Aligning predicates isn't easy, but crucially, it can be done without supervision and it's surprising that no one has followed in their steps. The authors are careful to distinguish the similarity-based graph clustering algorithm they use to align predicates from event coreference, claiming that the latter is more restricted in focusing on individual documents and exclusively on strict identity between events. While event identity is indeed too stringent a requirement for aligning predicates, numerous corpora have been released that generalize the traditional setting to include subevent and quasi-identity relations \citep{glavavs2014hieve, hong-etal-2016-building, o2016richer, gantt2021decomposing}, and these could potentially permit a similar augmentation technique, but one that relies on gold \textit{intra}-document relations between predicates, rather than silver \textit{cross}-document ones.

\subsection{Model Design and Evaluation Problems}
\label{subsec:model_eval_problems}
Data issues aside, the models and evaluation methods considered in \S\ref{sec:arg_linking} reveal various design problems that future ones should avoid. The issues that follow are presented in no particular order.

\subsubsection{Robustness to trigger-argument distance} If there is one kind of performative consistency argument linking systems should demonstrate, it is consistency across trigger-argument distances. This is especially true given the restricted context windows universally employed by the systems surveyed here. If system performance degrades precipitously as one moves away from the trigger sentence, it suggests that the system is really just a model for SRL that incidentally does argument linking---and poorly.

Historically, researchers have only very infrequently reported performance breakdowns by sentence distance, which is inexcusable. Fortunately, the trend seems to have reversed in work on RAMS. The two span-based models \citep{ebner-etal-2020-multi, chen2020joint} report encouraging results on this front (see Table \ref{tab:sent_dist}, left side). This is perhaps unsurprising, considering that there really is nothing in the design of either model that suggests an over-reliance on syntactic cues or other sentence-level features. In contrast, the head word-based model from \citet{zhang2020two} (Table \ref{tab:sent_dist}, right side) shows dramatic drops in performance as the distance between trigger and argument grows. While it should be acknowledged that \citeauthor{zhang2020two} have tackled a harder problem---full span detection and linking---than either of the other attempts on RAMS, the finding is still disappointing. It may be that argument head words by themselves provide insufficient signal for linking when not in the trigger sentence, and that the competitive performance \citeauthor{zhang2020two} achieve is driven by strong SRL alone.

\begin{table}[]
    \centering
    \begin{tabular}{c|cc}
    \toprule
         Dist & \citeauthor{ebner-etal-2020-multi} (TCD) & \citeauthor{chen2020joint}  \\
    \midrule
         -2 & 75.7 & 77.2 \\
         -1 & 73.7 & 74.4 \\
         0 & 75.0 & 79.6 \\
         1 & 76.5 & 77.0 \\
         2 & 79.1 & 78.7 \\
    \bottomrule
    \end{tabular}
    \quad
    \begin{tabular}{c|c}
    \toprule
    Dist & \citeauthor{zhang2020two} (TCD) \\
    \midrule
         -2 & 15.6 \\
         -1 & 15.3 \\
         0 & 43.4 \\
         1 & 17.8 \\
         2 & 8.5 \\
    \bottomrule
    \end{tabular}
    \caption{\textbf{Left}: Dev F1 by sentence distance from trigger for \citet{ebner-etal-2020-multi} and \citet{chen2020joint}. The evaluation setting assumes gold argument spans. \textbf{Right}: Dev \textit{span} F1 by sentence distance for \citet{zhang2020two}. The evaluation setting does \textit{not} assume gold argument spans, which accounts for the lower scores.}
    \label{tab:sent_dist}
\end{table}

\subsubsection{Getting away from gold spans} Linking gold argument spans is not a real task. It is a useful diagnostic to evaluate upper bounds on model performance, but gold arguments do not exist in the wild. Accordingly, the primary benchmarks against which argument linking systems are evaluated should not be ones that assume gold spans. It is, of course, acceptable to propose a pipelined \textit{model} in which a linking-only component takes argument spans from an upstream component, as was frequently done for task 2 on SemEval. I am arguing only that argument linking, properly conceived, \textit{includes} span detection and so should be evaluated on that basis.

\subsubsection{Beyond context windows} Argument linking, as it has been realized in the models presented in this survey, is not truly a document-level problem; it is a multi-sentence problem, owing to the assumption of a context window. From a resource development perspective, context windows are perfectly reasonable, as it is usually too demanding to ask annotators to consider an entire document when labeling each role of an event trigger. And even from the perspective of the problem itself, it makes sense to concentrate energies on sentences closer to the trigger, as there are diminishing returns on computational investment as one moves further away from it \citep{gerber2010beyond}. But with few exceptions, context windows seem to have discouraged researchers from \textit{thinking} at the document level and so they largely have not modeled document structure where it may be useful to do so. There are numerous ways to go about this. At one extreme, on can assume an ontology of document-level templates or scripts with roles that must be filled, in which case argument linking effectively becomes \textit{role-filler entity extraction (REE)}, a component of template filling \citep{sundheim1992overview, li2021document}. This has its place---particularly in restricted domains where only a handful of kinds of scenarios are likely to crop up---but it clearly doesn't generalize.

What would it mean to instead \textit{learn} document structure as part of argument linking? One way to cache this out would be as joint SRL and coreference. We have seen pipelined models that use existing tools to obtain silver SRL and coreference data, but \citet{o2019bringing} argues that \textit{end-to-end} neural models trained jointly on both tasks are perhaps the most promising way forward for argument linking, and I am inclined to agree. He observes that, increasingly, neural models for SRL and for coreference have very similar architectures. At base, most operate by comparing span representations. In both cases, an initial set of candidate spans is pruned. For coreference models, the spans that remain after pruning are candidate coreferent mentions for a target mention, and some scoring function is used to rank the candidates and select the one most likely to be coreferent with the target. For SRL, the remaining spans are candidate arguments for a predicate, and one or more scoring functions is used to determine whether the candidate is an argument of the predicate and, if so, what role it fills.

\citeauthor{o2019bringing}'s proposal for combining the two entails treating roles themselves as additional ``mentions'' that can be scored relative to other arguments. Thus, for a sentence like \textit{John ate his bagel}, the model would consider not only the argument spans ``John,'' ``bagel,'' and ``his bagel'' but the role spans ``arg0 of ate'' and ``arg1 of ate,'' and would hopefully learn coreference clusters \{John, his, arg0 of ate\} and \{bagel, arg1 of ate\}. Clustering the semantic roles with the arguments in this way would thus allow arguments to be linked across sentence boundaries. The major difficulty of such a model is the same one that faces all span-based ones---namely, how to reduce the quadratic number of candidates.

Nonetheless, it seems a promising avenue for further exploration and to my knowledge no one has attempted it. In fact, the only effort to do joint coreference and SRL at all appears to be \citet{aralikatte2020joint}, but they do not treat roles as mentions as \citeauthor{o2019bringing} suggests and are interested in the benefits that joint training can confer on the two tasks individually, rather than in exploiting them for argument linking. However, they achieve impressive results on both, which is some reason to think \citeauthor{o2019bringing}'s proposal just might work.

\newpage

\bibliography{references}

\begin{thebibliography}{72}
\providecommand{\natexlab}[1]{#1}
\providecommand{\url}[1]{\texttt{#1}}
\expandafter\ifx\csname urlstyle\endcsname\relax
  \providecommand{\doi}[1]{doi: #1}\else
  \providecommand{\doi}{doi: \begingroup \urlstyle{rm}\Url}\fi

\bibitem[Aralikatte et~al.(2020)Aralikatte, Abdou, Lent, Hershcovich, and
  S{\o}gaard]{aralikatte2020joint}
R.~Aralikatte, M.~Abdou, H.~Lent, D.~Hershcovich, and A.~S{\o}gaard.
\newblock Joint semantic analysis with document-level cross-task coherence
  rewards.
\newblock \emph{arXiv preprint arXiv:2010.05567}, 2020.

\bibitem[Baker et~al.(1998)Baker, Fillmore, and Lowe]{baker1998berkeley}
C.~F. Baker, C.~J. Fillmore, and J.~B. Lowe.
\newblock The berkeley framenet project.
\newblock In \emph{36th Annual Meeting of the Association for Computational
  Linguistics and 17th International Conference on Computational Linguistics,
  Volume 1}, pages 86--90, 1998.

\bibitem[Banarescu et~al.(2013)Banarescu, Bonial, Cai, Georgescu, Griffitt,
  Hermjakob, Knight, Koehn, Palmer, and Schneider]{banarescu2013abstract}
L.~Banarescu, C.~Bonial, S.~Cai, M.~Georgescu, K.~Griffitt, U.~Hermjakob,
  K.~Knight, P.~Koehn, M.~Palmer, and N.~Schneider.
\newblock Abstract meaning representation for sembanking.
\newblock In \emph{Proceedings of the 7th Linguistic Annotation Workshop and
  Interoperability with Discourse}, pages 178--186, 2013.

\bibitem[Bohnet(2010)]{bohnet2010top}
B.~Bohnet.
\newblock Top accuracy and fast dependency parsing is not a contradiction.
\newblock In \emph{Proceedings of the 23rd International Conference on
  Computational Linguistics (COLING 2010)}, pages 89--97, 2010.

\bibitem[Cardie(2020)]{cardie2020emnlp}
C.~Cardie.
\newblock Information extraction through the years: How did we get here?
\newblock EMNLP keynote, 2020.
\newblock URL
  \url{https://slideslive.com/38938634/information-extraction-through-the-years-how-did-we-get-here}.

\bibitem[Chen et~al.(2010)Chen, Schneider, Das, and Smith]{chen2010semafor}
D.~Chen, N.~Schneider, D.~Das, and N.~A. Smith.
\newblock {SEMAFOR}: Frame argument resolution with log-linear models.
\newblock In \emph{Proceedings of the 5th international workshop on semantic
  evaluation}, pages 264--267, 2010.

\bibitem[Chen et~al.(2020)Chen, Chen, and Van~Durme]{chen2020joint}
Y.~Chen, T.~Chen, and B.~Van~Durme.
\newblock Joint modeling of arguments for event understanding.
\newblock In \emph{Proceedings of the First Workshop on Computational
  Approaches to Discourse}, pages 96--101, 2020.

\bibitem[Cheng and Erk(2018)]{cheng2018implicitevent}
P.~Cheng and K.~Erk.
\newblock Implicit argument prediction with event knowledge.
\newblock In \emph{Proceedings of the 2018 Conference of the North {A}merican
  Chapter of the Association for Computational Linguistics: Human Language
  Technologies, Volume 1 (Long Papers)}, pages 831--840, New Orleans,
  Louisiana, June 2018. Association for Computational Linguistics.
\newblock \doi{10.18653/v1/N18-1076}.
\newblock URL \url{https://www.aclweb.org/anthology/N18-1076}.

\bibitem[Cheng and Erk(2019)]{cheng2019implicitreading}
P.~Cheng and K.~Erk.
\newblock Implicit argument prediction as reading comprehension.
\newblock In \emph{Proceedings of the AAAI Conference on Artificial
  Intelligence}, volume~33, pages 6284--6291, 2019.

\bibitem[Chincnor and Sundheim(2003)]{muc6}
N.~Chincnor and B.~Sundheim.
\newblock Message understanding conference ({MUC}) 6.
\newblock 2003.
\newblock \doi{10.35111/wbcc-y063}.

\bibitem[Cho et~al.(2014)Cho, Van~Merri{\"e}nboer, Gulcehre, Bahdanau,
  Bougares, Schwenk, and Bengio]{cho2014learning}
K.~Cho, B.~Van~Merri{\"e}nboer, C.~Gulcehre, D.~Bahdanau, F.~Bougares,
  H.~Schwenk, and Y.~Bengio.
\newblock Learning phrase representations using rnn encoder-decoder for
  statistical machine translation.
\newblock In \emph{Empirical Methods in Natural Language Processing (EMNLP
  2014)}, 2014.

\bibitem[Das et~al.(2010{\natexlab{a}})Das, Schneider, Chen, and
  Smith]{das2010probabilistic}
D.~Das, N.~Schneider, D.~Chen, and N.~A. Smith.
\newblock Probabilistic frame-semantic parsing.
\newblock In \emph{Human language technologies: The 2010 Annual Conference of
  the North American chapter of the Association for Computational Linguistics
  (NAACL-2010)}, pages 948--956, 2010{\natexlab{a}}.

\bibitem[Das et~al.(2010{\natexlab{b}})Das, Schneider, Chen, and
  Smith]{das2010semafor}
D.~Das, N.~Schneider, D.~Chen, and N.~A. Smith.
\newblock {SEMAFOR} 1.0: A probabilistic frame-semantic parser.
\newblock \emph{Language Technologies Institute, School of Computer Science,
  Carnegie Mellon University}, 2010{\natexlab{b}}.

\bibitem[De~Smet et~al.(2005)]{de2005corpus}
H.~De~Smet et~al.
\newblock A corpus of late modern english texts.
\newblock \emph{ICAME journal}, 29\penalty0 (29):\penalty0 69--82, 2005.

\bibitem[Devlin et~al.(2019)Devlin, Chang, Lee, and Toutanova]{devlin2019bert}
J.~Devlin, M.-W. Chang, K.~Lee, and K.~Toutanova.
\newblock {BERT}: Pre-training of deep bidirectional transformers for language
  understanding.
\newblock In \emph{Proceedings of the 2019 Conference of the North American
  Chapter of the Association for Computational Linguistics: Human Language
  Technologies}, pages 4171--4186, 2019.

\bibitem[Dowty(1991)]{dowty1991thematic}
D.~Dowty.
\newblock Thematic proto-roles and argument selection.
\newblock \emph{Language}, 67\penalty0 (3):\penalty0 547--619, 1991.

\bibitem[Dozat and Manning(2017)]{dozat2016deep}
T.~Dozat and C.~D. Manning.
\newblock Deep biaffine attention for neural dependency parsing.
\newblock In \emph{International Conference on Learning Representations}, 2017.

\bibitem[Ebner et~al.(2020)Ebner, Xia, Culkin, Rawlins, and
  Van~Durme]{ebner-etal-2020-multi}
S.~Ebner, P.~Xia, R.~Culkin, K.~Rawlins, and B.~Van~Durme.
\newblock Multi-sentence argument linking.
\newblock In \emph{Proceedings of the 58th Annual Meeting of the Association
  for Computational Linguistics}, pages 8057--8077, Online, July 2020.
  Association for Computational Linguistics.
\newblock \doi{10.18653/v1/2020.acl-main.718}.
\newblock URL \url{https://www.aclweb.org/anthology/2020.acl-main.718}.

\bibitem[Feizabadi(2019)]{feizabadi2019strategies}
P.~S. Feizabadi.
\newblock \emph{Strategies to Address Data Sparseness in Implicit Semantic Role
  Labeling}.
\newblock PhD thesis, Ruprecht-Karls-Universit¨at Heidelberg, 2019.

\bibitem[Feizabadi and Pad{\'o}(2015)]{feizabadi2015combining}
P.~S. Feizabadi and S.~Pad{\'o}.
\newblock Combining seemingly incompatible corpora for implicit semantic role
  labeling.
\newblock In \emph{Proceedings of the Fourth Joint Conference on Lexical and
  Computational Semantics}, pages 40--50, 2015.

\bibitem[Fillmore et~al.(1976)]{fillmore1976frame}
C.~J. Fillmore et~al.
\newblock Frame semantics and the nature of language.
\newblock In \emph{Annals of the New York Academy of Sciences: Conference on
  the Origin and Development of Language and Speech}, volume 280, pages 20--32.
  New York, 1976.

\bibitem[Gantt et~al.(2021)Gantt, Glass, and White]{gantt2021decomposing}
W.~Gantt, L.~Glass, and A.~S. White.
\newblock Decomposing and recomposing event structure.
\newblock \emph{arXiv preprint arXiv:2103.10387}, 2021.

\bibitem[Gerber and Chai(2010)]{gerber2010beyond}
M.~Gerber and J.~Chai.
\newblock Beyond nombank: A study of implicit arguments for nominal predicates.
\newblock In \emph{Proceedings of the 48th Annual Meeting of the Association
  for Computational Linguistics}, pages 1583--1592, 2010.

\bibitem[Gerber and Chai(2012)]{gerber2012semantic}
M.~Gerber and J.~Y. Chai.
\newblock Semantic role labeling of implicit arguments for nominal predicates.
\newblock \emph{Computational Linguistics}, 38\penalty0 (4):\penalty0 755--798,
  2012.

\bibitem[Gildea and Jurafsky(2002)]{gildea2002automatic}
D.~Gildea and D.~Jurafsky.
\newblock Automatic labeling of semantic roles.
\newblock \emph{Computational Linguistics}, 28\penalty0 (3):\penalty0 245--288,
  2002.

\bibitem[Glava{\v{s}} et~al.(2014)Glava{\v{s}}, {\v{S}}najder, Kordjamshidi,
  and Moens]{glavavs2014hieve}
G.~Glava{\v{s}}, J.~{\v{S}}najder, P.~Kordjamshidi, and M.-F. Moens.
\newblock Hieve: A corpus for extracting event hierarchies from news stories.
\newblock In \emph{Proceedings of 9th Language Resources and Evaluation
  Conference}, pages 3678--3683. ELRA, 2014.

\bibitem[Graff and Cieri(2003)]{gigaword}
D.~Graff and C.~Cieri.
\newblock English {G}igaword.
\newblock 2003.
\newblock \doi{10.35111/kcqk-v224}.

\bibitem[He et~al.(2017)He, Lee, Lewis, and Zettlemoyer]{he-etal-2017-deep}
L.~He, K.~Lee, M.~Lewis, and L.~Zettlemoyer.
\newblock Deep semantic role labeling: What works and what{'}s next.
\newblock In \emph{Proceedings of the 55th Annual Meeting of the Association
  for Computational Linguistics (Volume 1: Long Papers)}, pages 473--483,
  Vancouver, Canada, July 2017. Association for Computational Linguistics.
\newblock \doi{10.18653/v1/P17-1044}.
\newblock URL \url{https://www.aclweb.org/anthology/P17-1044}.

\bibitem[He et~al.(2018)He, Lee, Levy, and Zettlemoyer]{he-etal-2018-jointly}
L.~He, K.~Lee, O.~Levy, and L.~Zettlemoyer.
\newblock Jointly predicting predicates and arguments in neural semantic role
  labeling.
\newblock In \emph{Proceedings of the 56th Annual Meeting of the Association
  for Computational Linguistics (Volume 2: Short Papers)}, pages 364--369,
  Melbourne, Australia, July 2018. Association for Computational Linguistics.
\newblock \doi{10.18653/v1/P18-2058}.
\newblock URL \url{https://www.aclweb.org/anthology/P18-2058}.

\bibitem[Hermann et~al.(2015)Hermann, Ko{\v{c}}isk{\`y}, Grefenstette,
  Espeholt, Kay, Suleyman, and Blunsom]{hermann2015teaching}
K.~M. Hermann, T.~Ko{\v{c}}isk{\`y}, E.~Grefenstette, L.~Espeholt, W.~Kay,
  M.~Suleyman, and P.~Blunsom.
\newblock Teaching machines to read and comprehend.
\newblock \emph{Advances in Neural Information Processing Systems}, pages
  1693--1701, 2015.

\bibitem[Hong et~al.(2016)Hong, Zhang, O{'}Gorman, Horowit-Hendler, Ji, and
  Palmer]{hong-etal-2016-building}
Y.~Hong, T.~Zhang, T.~O{'}Gorman, S.~Horowit-Hendler, H.~Ji, and M.~Palmer.
\newblock Building a cross-document event-event relation corpus.
\newblock In \emph{Proceedings of the 10th Linguistic Annotation Workshop
  ({LAW}-X 2016)}, pages 1--6, Berlin, Germany, Aug. 2016. Association for
  Computational Linguistics.
\newblock \doi{10.18653/v1/W16-1701}.
\newblock URL \url{https://www.aclweb.org/anthology/W16-1701}.

\bibitem[Hovy et~al.(2006)Hovy, Marcus, Palmer, Ramshaw, and
  Weischedel]{hovy2006ontonotes}
E.~Hovy, M.~Marcus, M.~Palmer, L.~Ramshaw, and R.~Weischedel.
\newblock Ontonotes: the 90\% solution.
\newblock In \emph{Proceedings of the human language technology conference of
  the NAACL, Companion Volume: Short Papers}, pages 57--60, 2006.

\bibitem[Jackendoff(1987)]{jackendoff1987status}
R.~Jackendoff.
\newblock The status of thematic relations in linguistic theory.
\newblock \emph{Linguistic Inquiry}, 18\penalty0 (3):\penalty0 369--411, 1987.

\bibitem[Koomen et~al.(2005)Koomen, Punyakanok, Roth, and
  Yih]{koomen2005generalized}
P.~Koomen, V.~Punyakanok, D.~Roth, and W.-t. Yih.
\newblock Generalized inference with multiple semantic role labeling systems.
\newblock In \emph{Proceedings of the Ninth Conference on Computational Natural
  Language Learning (CoNLL-2005)}, pages 181--184, 2005.

\bibitem[Laparra and Rigau(2012)]{laparra2012exploiting}
E.~Laparra and G.~Rigau.
\newblock Exploiting explicit annotations and semantic types for implicit
  argument resolution.
\newblock In \emph{2012 IEEE Sixth International Conference on Semantic
  Computing}, pages 75--78. IEEE, 2012.

\bibitem[Laparra and Rigau(2013)]{laparra2013impar}
E.~Laparra and G.~Rigau.
\newblock Impar: A deterministic algorithm for implicit semantic role
  labelling.
\newblock In \emph{Proceedings of the 51st Annual Meeting of the Association
  for Computational Linguistics (Volume 1: Long Papers)}, pages 1180--1189,
  2013.

\bibitem[Lappin and Leass(1994)]{lappin1994algorithm}
S.~Lappin and H.~J. Leass.
\newblock An algorithm for pronominal anaphora resolution.
\newblock \emph{Computational Linguistics}, 20\penalty0 (4):\penalty0 535--561,
  1994.

\bibitem[Lee et~al.(2013)Lee, Chang, Peirsman, Chambers, Surdeanu, and
  Jurafsky]{lee2013deterministic}
H.~Lee, A.~Chang, Y.~Peirsman, N.~Chambers, M.~Surdeanu, and D.~Jurafsky.
\newblock Deterministic coreference resolution based on entity-centric,
  precision-ranked rules.
\newblock \emph{Computational Linguistics}, 39\penalty0 (4):\penalty0 885--916,
  2013.

\bibitem[Lee et~al.(2017)Lee, He, Lewis, and Zettlemoyer]{lee-etal-2017-end}
K.~Lee, L.~He, M.~Lewis, and L.~Zettlemoyer.
\newblock End-to-end neural coreference resolution.
\newblock In \emph{Proceedings of the 2017 Conference on Empirical Methods in
  Natural Language Processing}, pages 188--197, Copenhagen, Denmark, Sept.
  2017. Association for Computational Linguistics.
\newblock \doi{10.18653/v1/D17-1018}.
\newblock URL \url{https://www.aclweb.org/anthology/D17-1018}.

\bibitem[Li et~al.(2021)Li, Ji, and Han]{li2021document}
S.~Li, H.~Ji, and J.~Han.
\newblock Document-level event argument extraction by conditional generation.
\newblock \emph{arXiv preprint arXiv:2104.05919}, 2021.

\bibitem[Manning et~al.(2014)Manning, Surdeanu, Bauer, Finkel, Bethard, and
  McClosky]{manning2014stanford}
C.~D. Manning, M.~Surdeanu, J.~Bauer, J.~R. Finkel, S.~Bethard, and
  D.~McClosky.
\newblock The stanford corenlp natural language processing toolkit.
\newblock In \emph{Proceedings of 52nd annual meeting of the association for
  computational linguistics: system demonstrations}, pages 55--60, 2014.

\bibitem[Marcus et~al.()Marcus, Palmer, Ramshaw, and Xue]{marcusontonotes}
R.~W. E. H.~M. Marcus, M.~Palmer, R.~B. S. P.~L. Ramshaw, and N.~Xue.
\newblock Onto{N}otes: A large training corpus for enhanced processing.

\bibitem[Meyers et~al.(2004)Meyers, Reeves, Macleod, Szekely, Zielinska, Young,
  and Grishman]{meyers2004nombank}
A.~Meyers, R.~Reeves, C.~Macleod, R.~Szekely, V.~Zielinska, B.~Young, and
  R.~Grishman.
\newblock The {N}om{B}ank project: An interim report.
\newblock In \emph{Proceedings of the Workshop on Frontiers in Corpus
  Annotation}, pages 24--31, 2004.

\bibitem[Mikolov et~al.(2013)Mikolov, Sutskever, Chen, Corrado, and
  Dean]{mikolov2013advances}
T.~Mikolov, I.~Sutskever, K.~Chen, G.~S. Corrado, and J.~Dean.
\newblock Distributed representations of words and phrases and their
  compositionality.
\newblock \emph{Advances in Neural Information Processing Systems}, pages
  3111--9, 2013.

\bibitem[Miller(1998)]{miller1998wordnet}
G.~A. Miller.
\newblock \emph{WordNet: An electronic lexical database}.
\newblock MIT press, 1998.

\bibitem[Mitchell et~al.(2003)Mitchell, Strassel, Przybocki, Davis, Doddington,
  Grishman, Meyers, Brunstein, Ferro, and Sundheim]{ace2}
A.~Mitchell, S.~Strassel, M.~Przybocki, J.~Davis, G.~R. Doddington,
  R.~Grishman, A.~Meyers, A.~Brunstein, L.~Ferro, and B.~Sundheim.
\newblock {ACE}-2 {V}ersion 1.0.
\newblock 2003.
\newblock \doi{10.35111/kcqk-v224}.

\bibitem[Moor et~al.(2013)Moor, Roth, and Frank]{moor2013predicate}
T.~Moor, M.~Roth, and A.~Frank.
\newblock Predicate-specific annotations for implicit role binding: Corpus
  annotation, data analysis and evaluation experiments.
\newblock In \emph{Proceedings of the 10th International Conference on
  Computational Semantics (IWCS 2013)--Short Papers}, pages 369--375, 2013.

\bibitem[O'Gorman(2019)]{o2019bringing}
T.~J. O'Gorman.
\newblock \emph{Bringing Together Computational and Linguistic Models of
  Implicit Role Interpretation}.
\newblock PhD thesis, University of Colorado at Boulder, 2019.

\bibitem[Ouchi et~al.(2018)Ouchi, Shindo, and Matsumoto]{ouchi-etal-2018-span}
H.~Ouchi, H.~Shindo, and Y.~Matsumoto.
\newblock A span selection model for semantic role labeling.
\newblock In \emph{Proceedings of the 2018 Conference on Empirical Methods in
  Natural Language Processing}, pages 1630--1642, Brussels, Belgium, Oct.-Nov.
  2018. Association for Computational Linguistics.
\newblock \doi{10.18653/v1/D18-1191}.
\newblock URL \url{https://www.aclweb.org/anthology/D18-1191}.

\bibitem[O’Gorman et~al.(2016)O’Gorman, Wright-Bettner, and
  Palmer]{o2016richer}
T.~O’Gorman, K.~Wright-Bettner, and M.~Palmer.
\newblock Richer event description: Integrating event coreference with
  temporal, causal and bridging annotation.
\newblock In \emph{Proceedings of the 2nd Workshop on Computing News Storylines
  (CNS 2016)}, pages 47--56, 2016.

\bibitem[O’Gorman et~al.(2018)O’Gorman, Regan, Griffitt, Hermjakob, Knight,
  and Palmer]{o2018amr}
T.~O’Gorman, M.~Regan, K.~Griffitt, U.~Hermjakob, K.~Knight, and M.~Palmer.
\newblock Amr beyond the sentence: the multi-sentence amr corpus.
\newblock In \emph{Proceedings of the 27th International Conference on
  Computational Linguistics}, pages 3693--3702, 2018.

\bibitem[Palmer(2009)]{palmer2009semlink}
M.~Palmer.
\newblock Semlink: Linking {P}rop{B}ank, {V}erb{N}et and {F}rame{N}et.
\newblock In \emph{Proceedings of the Generative Lexicon Conference}, pages
  9--15. GenLex-09, Pisa, Italy, 2009.

\bibitem[Palmer et~al.(2005)Palmer, Gildea, and
  Kingsbury]{palmer2005proposition}
M.~Palmer, D.~Gildea, and P.~Kingsbury.
\newblock The proposition bank: An annotated corpus of semantic roles.
\newblock \emph{Computational Linguistics}, 31\penalty0 (1):\penalty0 71--106,
  2005.

\bibitem[Palmer et~al.(1986)Palmer, Dahl, Schiffman, Hirschman, Linebarger, and
  Dowding]{palmer-etal-1986-recovering-implicit}
M.~S. Palmer, D.~A. Dahl, R.~J. Schiffman, L.~Hirschman, M.~Linebarger, and
  J.~Dowding.
\newblock Recovering implicit information.
\newblock In \emph{24th Annual Meeting of the Association for Computational
  Linguistics}, pages 10--19, New York, New York, USA, July 1986. Association
  for Computational Linguistics.
\newblock \doi{10.3115/981131.981135}.
\newblock URL \url{https://www.aclweb.org/anthology/P86-1004}.

\bibitem[Peters et~al.(2018)Peters, Neumann, Iyyer, Gardner, Clark, Lee, and
  Zettlemoyer]{peters-etal-2018-deep}
M.~Peters, M.~Neumann, M.~Iyyer, M.~Gardner, C.~Clark, K.~Lee, and
  L.~Zettlemoyer.
\newblock Deep contextualized word representations.
\newblock In \emph{Proceedings of the 2018 Conference of the North {A}merican
  Chapter of the Association for Computational Linguistics: Human Language
  Technologies, Volume 1 (Long Papers)}, pages 2227--2237, New Orleans,
  Louisiana, June 2018. Association for Computational Linguistics.
\newblock \doi{10.18653/v1/N18-1202}.
\newblock URL \url{https://www.aclweb.org/anthology/N18-1202}.

\bibitem[Pradhan et~al.(2005)Pradhan, Hacioglu, Krugler, Ward, Martin, and
  Jurafsky]{pradhan2005support}
S.~Pradhan, K.~Hacioglu, V.~Krugler, W.~Ward, J.~H. Martin, and D.~Jurafsky.
\newblock Support vector learning for semantic argument classification.
\newblock \emph{Machine Learning}, 60\penalty0 (1):\penalty0 11--39, 2005.

\bibitem[Pradhan et~al.(2011)Pradhan, Ramshaw, Marcus, Palmer, Weischedel, and
  Xue]{pradhan2011conll}
S.~Pradhan, L.~Ramshaw, M.~Marcus, M.~Palmer, R.~Weischedel, and N.~Xue.
\newblock Conll-2011 shared task: Modeling unrestricted coreference in
  ontonotes.
\newblock In \emph{Proceedings of the Fifteenth Conference on Computational
  Natural Language Learning: Shared Task}, pages 1--27, 2011.

\bibitem[Prasad et~al.(2008)Prasad, Dinesh, Lee, Miltsakaki, Robaldo, Joshi,
  and Webber]{prasad2008penn}
R.~Prasad, N.~Dinesh, A.~Lee, E.~Miltsakaki, L.~Robaldo, A.~Joshi, and
  B.~Webber.
\newblock The {P}enn {D}iscourse {T}ree{B}ank 2.0.
\newblock In \emph{Proceedings of the Sixth International Conference on
  Language Resources and Evaluation ({LREC}'08)}, Marrakech, Morocco, May 2008.
  European Language Resources Association (ELRA).
\newblock URL
  \url{http://www.lrec-conf.org/proceedings/lrec2008/pdf/754_paper.pdf}.

\bibitem[Roth and Frank(2012)]{roth2012aligning}
M.~Roth and A.~Frank.
\newblock Aligning predicate argument structures in monolingual comparable
  texts: A new corpus for a new task.
\newblock In \emph{* SEM 2012: The First Joint Conference on Lexical and
  Computational Semantics--Volume 1: Proceedings of the main conference and the
  shared task, and Volume 2: Proceedings of the Sixth International Workshop on
  Semantic Evaluation (SemEval 2012)}, pages 218--227, 2012.

\bibitem[Roth and Frank(2013)]{roth2013automatically}
M.~Roth and A.~Frank.
\newblock Automatically identifying implicit arguments to improve argument
  linking and coherence modeling.
\newblock In \emph{Second Joint Conference on Lexical and Computational
  Semantics (* SEM), Volume 1: Proceedings of the Main Conference and the
  Shared Task: Semantic Textual Similarity}, pages 306--316, 2013.

\bibitem[Ruppenhofer et~al.(2010)Ruppenhofer, Sporleder, Morante, Baker, and
  Palmer]{ruppenhofer-etal-2010-semeval}
J.~Ruppenhofer, C.~Sporleder, R.~Morante, C.~Baker, and M.~Palmer.
\newblock {S}em{E}val-2010 task 10: Linking events and their participants in
  discourse.
\newblock In \emph{Proceedings of the 5th International Workshop on Semantic
  Evaluation}, pages 45--50, Uppsala, Sweden, July 2010. Association for
  Computational Linguistics.
\newblock URL \url{https://www.aclweb.org/anthology/S10-1008}.

\bibitem[Schenk and Chiarcos(2016)]{schenk2016unsupervised}
N.~Schenk and C.~Chiarcos.
\newblock Unsupervised learning of prototypical fillers for implicit semantic
  role labeling.
\newblock In \emph{Proceedings of the 2016 Conference of the North American
  Chapter of the Association for Computational Linguistics: Human Language
  Technologies}, pages 1473--1479, 2016.

\bibitem[Schuler(2005)]{schuler2005verbnet}
K.~K. Schuler.
\newblock \emph{VerbNet: A broad-coverage, comprehensive verb lexicon}.
\newblock PhD thesis, University of Pennsylvania, 2005.

\bibitem[Silberer and Frank(2012)]{silberer2012casting}
C.~Silberer and A.~Frank.
\newblock Casting implicit role linking as an anaphora resolution task.
\newblock In \emph{Agirre E, Bos J, Diab M, Manandhar S, Marton Y, Yuret D. SEM
  2012: The First Joint Conference on Lexical and Computational Semantics. 2012
  Jun 7-8; Montr{\'e}al, Canada. Stroudsburg: ACL; 2012. p. 1-10.} ACL
  (Association for Computational Linguistics), 2012.

\bibitem[Sundheim(1992)]{sundheim1992overview}
B.~M. Sundheim.
\newblock Overview of the fourth {M}essage {U}nderstanding {E}valuation and
  {C}onference.
\newblock In \emph{{F}ourth {M}essage {U}understanding {C}onference ({MUC}-4):
  Proceedings of a Conference Held in {M}c{L}ean, {V}irginia, {J}une 16-18,
  1992}, 1992.
\newblock URL \url{https://www.aclweb.org/anthology/M92-1001}.

\bibitem[Surdeanu et~al.(2008)Surdeanu, Johansson, Meyers, M{\`a}rquez, and
  Nivre]{surdeanu2008conll}
M.~Surdeanu, R.~Johansson, A.~Meyers, L.~M{\`a}rquez, and J.~Nivre.
\newblock The {CoNLL} 2008 shared task on joint parsing of syntactic and
  semantic dependencies.
\newblock In \emph{CoNLL 2008: Proceedings of the Twelfth Conference on
  Computational Natural Language Learning}, pages 159--177, 2008.

\bibitem[Van~de Cruys(2014)]{van2014neural}
T.~Van~de Cruys.
\newblock A neural network approach to selectional preference acquisition.
\newblock In \emph{Empirical Methods in Natural Language Processing (EMNLP)},
  pages 26--35, 2014.

\bibitem[Vaswani et~al.(2017)Vaswani, Shazeer, Parmar, Uszkoreit, Jones, Gomez,
  Kaiser, and Polosukhin]{vaswani2017attention}
A.~Vaswani, N.~Shazeer, N.~Parmar, J.~Uszkoreit, L.~Jones, A.~N. Gomez,
  L.~Kaiser, and I.~Polosukhin.
\newblock Attention is all you need.
\newblock In \emph{Advances in Neural Information Processing Systems}, pages
  5998--6008, 2017.

\bibitem[Wechsler(1995)]{wechsler1995semantic}
S.~Wechsler.
\newblock \emph{The Semantic Basis of Argument Structure}.
\newblock Dissertations in Linguistics. Cambridge University Press, 1995.
\newblock ISBN 9781881526698.
\newblock URL \url{https://books.google.com/books?id=t-rotAEACAAJ}.

\bibitem[Williams(2015)]{williams2015arguments}
A.~Williams.
\newblock \emph{Arguments in syntax and semantics}.
\newblock Cambridge University Press, 2015.

\bibitem[Xue and Palmer(2004)]{xue-palmer-2004-calibrating}
N.~Xue and M.~Palmer.
\newblock Calibrating features for semantic role labeling.
\newblock In \emph{Proceedings of the 2004 Conference on Empirical Methods in
  Natural Language Processing}, pages 88--94, Barcelona, Spain, July 2004.
  Association for Computational Linguistics.
\newblock URL \url{https://www.aclweb.org/anthology/W04-3212}.

\bibitem[Zhang et~al.(2020)Zhang, Kong, Liu, Ma, and Hovy]{zhang2020two}
Z.~Zhang, X.~Kong, Z.~Liu, X.~Ma, and E.~Hovy.
\newblock A two-step approach for implicit event argument detection.
\newblock In \emph{Proceedings of the 58th Annual Meeting of the Association
  for Computational Linguistics}, pages 7479--7485, 2020.

\end{thebibliography}
\bibliographystyle{abbrvnat}

\end{document}